\documentclass{article}
\pdfoutput=1
\usepackage{arxiv}
\usepackage{hyperref}
\hypersetup{
  pdfinfo={
    Title={title},
    Author={author},
  }
}
\usepackage{pdfpages}
\usepackage[utf8]{inputenc} 
\usepackage[T1]{fontenc}    
\usepackage{hyperref}       
\usepackage{url}            
\usepackage{booktabs}       
\usepackage{amsfonts}       
\usepackage{nicefrac}       
\usepackage{microtype}      
\usepackage{lipsum}
\usepackage{graphicx}
\graphicspath{ {./images/} }
\usepackage{times}  
\usepackage{helvet} 
\usepackage{courier}  
\usepackage{graphicx} 
\usepackage{caption}
\usepackage{tcolorbox}
\usepackage{epsfig}
\usepackage{amsmath}
\usepackage{fancyref}
\usepackage{amsthm}
\usepackage[ruled,linesnumbered]{algorithm2e}
\usepackage{subfigure}
\usepackage{algpseudocode}
\usepackage{enumerate}
\usepackage{epstopdf}
\newtheorem{theorem}{Theorem}[section]
\newtheorem{definition}{Definition}[section]
\newtheorem{corollary}{Corollary}[section]
\newtheorem*{corollary-non}{Corollary}
\newtheorem*{theorem-non}{Theorem}
\newtheorem{lemma}{Lemma}[section]
\newtheorem*{lemma-non}{Lemma}

\title{A Variance Controlled Stochastic Method with Biased Estimation for Faster Non-convex Optimization}

\author{
 Jia Bi \\
  School of Electronic and Computer Science\\
  University of Southampton\\
  Southampton, United Kingdom \\
  \texttt{J.Bi@soton.ac.uk} \\
   \And
 Steve R. Gunn \\
  School of Electronic and Computer Science\\
  University of Southampton\\
  Southampton, United Kingdom \\
  \texttt{srg@soton.ac.uk} \\
}

\begin{document}
\maketitle
\begin{abstract}
 In this paper, we proposed a new technique, {\em variance controlled stochastic gradient} (VCSG), to improve the performance of the stochastic variance reduced gradient (SVRG) algorithm. To avoid over-reducing the variance of gradient by SVRG, a hyper-parameter $\lambda$ is introduced in VCSG that is able to control the reduced variance of SVRG. Theory shows that the optimization method can converge by using an unbiased gradient estimator, but in practice, biased gradient estimation can allow more efficient convergence to the vicinity since an unbiased approach is computationally more expensive. $\lambda$ also has the effect of balancing the trade-off between unbiased and biased estimations. Secondly, to minimize the number of full gradient calculations in SVRG, a variance-bounded batch is introduced to reduce the number of gradient calculations required in each iteration. For smooth non-convex functions, the proposed algorithm converges to an approximate first-order stationary point (i.e. $\mathbb{E}\|\nabla{f}(x)\|^{2}\leq\epsilon$) within $\mathcal{O}(min\{1/\epsilon^{3/2},n^{1/4}/\epsilon\})$ number of stochastic gradient evaluations, which improves the leading gradient complexity of stochastic gradient-based method SCSG~\cite{NIPS2017_6829} $(\mathcal{O}(min\{1/\epsilon^{5/3},n^{2/3}/\epsilon\})$. It is shown theoretically and experimentally that VCSG can be deployed to improve convergence.
\end{abstract}





\maketitle

\section{Introduction}
We study smooth non-convex optimization problems which is shown in~Eq.\ref{main-problem}, 
\begin{equation}
\min_{x\in\mathbb{R}^{d}}f(x),\ \  f(x):=\frac{1}{n}\sum^{n}_{i=1}f_{i}(x),
\label{main-problem}
\end{equation}
where each component $f_{i}(x)(i\in[n])$ is possibly non-convex and Lipschitz ($\mathcal{L}-smooth$)~\cite{Strongin:2000:GON:1200319,Reddi:2016:SVR:3045390.3045425}. We use $\mathcal{F}_{n}$ to denote all $f_{i}(x)$ functions of the form in Eq.~\ref{main-problem}, and optimize such functions using Incremental First-order (IFO) and Stochastic First-Order (SFO) Oracles, which are defined in Definition~\ref{ifo} and~\ref{sfo} respectively.
\begin{definition}~\cite{agarwal-bottou-2015}
	For a function $F(x)=\dfrac{1}{n}\sum_{i}f_{i}(x)$, an IFO takes an index $i\in[n]$ and a point $x\in\mathbb{R}^{d}$, and returns the pair $\left(f_{i}(x),\nabla{f}_{i}(x)\right)$.
	\label{ifo}
\end{definition}
\begin{definition}~\cite{pmlr-v80-qu18a}
	For a function $F(x)=\mathbb{E}_{y}f(x,y)$ where $y\sim P$, a SFO returns the stochastic gradient $G(x_{k},y_{k})=\nabla_{x}f(x_{k},y_{k})$ where $y_{k}$ is a sample drawn i.i.d. from $P$ in the $k_{th}$ call.
	\label{sfo}
\end{definition}
Non-convex optimization is required for many statistical learning tasks ranging from generalized linear models to deep neural networks~\cite{McCullagh:1989,NIPS2017_6829}. 
Many earlier works have focused on the asymptotic performance of algorithms~\cite{doi:10.1080/10556789408805582,doi:10.1137/S1052623495287022,doi:10.1137/S1052623495294797} and non-asymptotic complexity bounds have emerged~\cite{NIPS2017_6829}. To our knowledge, the first non-asymptotic convergence for stochastic gradient descent (SGD) was proposed by~\cite{DBLP:journals/mp/GhadimiL16} with $\mathcal{O}(1/\epsilon^{2})$. Full batch gradient descent (GD) is known to ensure convergence with $\mathcal{O}(n/\epsilon)$. Compared with SGD, GD's rate has better dependence on $\epsilon$ but worse dependence on $n$ due to the requirement of computing a full gradient. Variance reduced (VR) methods based on SGD, e.g. Stochastic Variance Reduced Gradient (SVRG)~\cite{NIPS2013_4937}, SAGA~\cite{DBLP:journals/corr/DefazioBL14} have been shown to achieve better dependence on $n$ than GD on non-convex problems with $\mathcal{O}(n+(n^{2/3}/\epsilon))$~\cite{7798553,Reddi:2016:SVR:3045390.3045425}. However, compared with SGD, the rate of VR based methods still have worse dependence on $\epsilon$ unless $\epsilon\ll n^{-2/3}$. Recently, \cite{NIPS2017_6829} proposed a method called SCSG combining the benefits of SGD and SVRG, which is the first algorithm that achieves a better rate than SGD and is no worse than SVRG with $\mathcal{O}(1/\epsilon^{5/3}\wedge n^{2/3}/\epsilon)$\footnote{$(a\wedge b)$ means $\min(a,b)$}. SNVRG proposed by~\cite{NIPS2018_7648} uses nested variance reduction to reduce the result of SCSG to $\tilde{\mathcal{O}}((1/\epsilon^{3/2})\wedge (n^{1/2}/\epsilon))$\footnote{$\tilde{\mathcal{O}}(\cdot)$ hides the logarithmic factors} that outperforms both SGD, GD and SVRG. Further SPIDER~\cite{NIPS2018_7349} proposes their both lower and upper bound as $\mathcal{O}(1/\epsilon^{3/2}\wedge n^{1/2}/\epsilon)$. \cite{DBLP:journals/corr/abs-1912-02365} provide the lower bound of $\epsilon$-based convergence rate as $\mathcal{O}(1/\epsilon^{3/2})$ which verify the results of SPIDER. As a result, the $\epsilon$-related convergence rate $\mathcal{O}(1/\epsilon^{3/2})$ is likely to be the best currently. To the best of our knowledge, SPIDER is a leading result of gradient complexity for smooth non-convex optimization. Although SPIDER use averaged L-Lipschitz gradients, which slightly unfair to compare with many other results that use L-Lipschitz gradients, their work motivates the research question about whether an algorithm based on SGD and VR-based methods can further reduce the rate of SPIDER when it depends on $\epsilon$ in the regime of modest target accuracy and depends on $n$ in the regime of high target accuracy.

However, for SGD and VR-based stochastic algorithms, there still exists three challenges. Firstly, they do not require a full gradient computation as in the SVRG method. As a result, SCSG, SNVRG, SPIDER reduce the full batch-size from $\mathcal{O}(n)$ to its subset as $\mathcal{O}(B)$ where $B<n$, which can significantly reduce the computational cost. However, it is challenging to appropriately scale the subset of samples in each stage of optimization to accelerate the convergence and achieve the same accuracy with full samples. Secondly, the variance of SGD is reduced by VR methods since the gradient of SGD is often too noisy to converge. However, VR schemes reduce the ability to escape local minima in later iterations due to a diminishing variance~\cite{10.1007/978-3-030-29911-8_26}. The challenge of SGD and VR methods is, therefore, to control the variance of gradients. Lastly, there exists a trade-off between biased/unbiased estimation in VR-based algorithms. SVRG is an unbiased estimation that can guarantee to converge but is not efficient to be used in real-world applications. Biased estimation can give a lower upper bound of the mean squared error (MSE) loss function~\cite{DBLP:conf/nips/LiangBBJ09}, and many works have proposed asymptotically biased optimization with biased gradient estimators as an economical alternative to an unbiased version. These do not converge to the minima, but to their vicinity~\cite{CHEN1987217,DBLP:journals/corr/abs-1807-11880,DBLP:journals/corr/abs-1801-10247,doi:10.1080/17442508908833545}. These methods provide a good insight into the biased gradient search. However, they hold under restrictive conditions, which are very hard to verify for complex stochastic gradient algorithms. Thus, the last challenge is how to balance the unbiased and biased estimator in different stages of the non-convex optimization process. 

To address these three challenges, we propose our method Variance Controlled Stochastic Gradient(VCSG) which can control the reduced variance of the subset of gradients and choose the biased or unbiased estimator in each iteration to accelerate the convergence rate of non-convex optimization. The standard goal of non-convex optimization with provable guarantee is to estimate~\textit{approximate} local optima since finding global optimum by bounding technique is still a NP-hard~\cite{Agarwal2016FindingAL,NIPS2018_7533}. Alternatively, our algorithm can fast and~\textit{heuristic} turning a relatively good local optima into a global one. We summarize and list our main contributions:
\begin{itemize}
	\item We provide a new method VCSG, a well-balanced VR method for SGD to achieve a competitive convergence rate. We provide a theoretical analysis of our algorithm on non-convex problems. To the best of our knowledge, we provide the first analysis that the controlled variance reduction can achieve comparable or faster convergence than gradient-based optimization. Table~\ref{table6.1} compares the theoretical rates of convergence of five methods, which shows that VCSG has the faster rate of convergence than other methods. Here, we did not compare our result to SNVRG and SPIDER since both of their results are under averaged Lipschitz assumption, which is not same with our problem domain. We then show empirically that VCSG has faster rates of convergence than SGD, SVRG and SCSG. 
\begin{table*}[!ht]	
	\centering
	\addtolength{\tabcolsep}{1pt}
	\caption{Comparison of results on SFO ~Definition~\ref{sfo} and IFO calls ~Definition~\ref{ifo} of gradient methods for smooth non-convex problems. Full batch optimizations include GD~\cite{Nesterov:2014:ILC:2670022}, SGD~\cite{DBLP:journals/mp/GhadimiL16} and SVRG~\cite{ Reddi:2016:SVR:3045390.3045425,DBLP:conf/icml/ZhuH16}. Batch methods, e.g. SCSG~\cite{NIPS2017_6829}, 
	and VCSG use a subset of the full samples $n$ which can significantly reduce the computational complexity.}
	\label{table6.1}
	\SetAlgoLined
	{\def\arraystretch{0.1}
		\begin{tabular}{|c|c|c|c|c|}
			\hline
			Algorithms & SFO/IFO calls on Non-convex                                  
			& Batch size $B$    & Learning rate $\eta$              \\ \hline
			GD       & $\mathcal{O}(n/\epsilon)$             & $n$   & $\mathcal{O}(L^{-1})$   \\ \hline
			SGD       & $\mathcal{O}(1/\epsilon^{2})$        & $n$    & $\mathcal{O}(L^{-1})$  \\ \hline
			SVRG      & $\mathcal{O}(n+(n^{2/3}/\epsilon))$  & $n$    & $\mathcal{O}(L^{-1}n^{-2/3})$ \\ \hline
			SCSG     & $\mathcal{O}(1/\epsilon^{5/3}\wedge n^{2/3}/\epsilon)$   & $B (B<n)$ & $\mathcal{O}(L^{-1}(n^{-2/3}\wedge\epsilon^{4/3}))$ \\ \hline
			VCSG     & $\mathbf{\mathcal{O}(1/\epsilon^{3/2}\wedge n^{1/4}/\epsilon)}$ & $B (B<n)$ & $\mathcal{O}(L^{-1}\wedge L^{-1}B^{-1/2})$ \\ \hline
		\end{tabular}
	}
\end{table*}
	\item VCSG provides an appropriate sample size in each iteration by the controlled variance reduction, which can significantly save computational cost. 
	\item VCSG balances the trade-off in biased and unbiased estimation, which provides a fast convergence rate.
\end{itemize}
\section{Preliminaries}
We use $\|\cdot\|$ to denote the Euclidean norm for brevity throughout the paper. For our analysis, the background that are required to introduce definitions for $L$-smooth and $\epsilon$-accuracy which now are defined in~Definition~\ref{lsmooth1} and~Definition~\ref{d3} respectively.
\begin{definition}
     Assume the individual functions $f_{i}$ in~Eq.\ref{main-problem} are $\mathcal{L}$-smooth if there is a constant $L$ such that
	\begin{equation}
	\|\nabla f_{i}(x)-\nabla f_{i}(y)\|\leq L\| x-y\|,\forall x,y \in \mathbb{R}^{d}
	\end{equation} 
	for some $L<\infty$ and for all $i\in\{1,...,n\}$.
\label{lsmooth1}
\end{definition}
We analyze convergence rates for Eq.\ref{main-problem} and apply $\|\nabla{f}\|^{2}\leq\epsilon$ convergence criterion by~\cite{articleNesterov}, which the concept of $\epsilon$-$\textit{accurate}$ is defined in Definition~\ref{d3}. Moreover, the minimum IFO/SFO in Definition~\ref{ifo} and ~\ref{sfo} to reach an $\epsilon-$accurate solution is denoted by $C_{comp}(\epsilon)$, and its complexity bound is denoted by $\mathbb{E}C_{comp}(\epsilon)$.
\begin{definition}
	\label{d3}
A point $x$ is called $\epsilon$-accurate if $\|\nabla f(x)\|^{2}\leq\epsilon$. An iterative stochastic algorithm can achieve $\epsilon$-accuracy within $t$ iterations if $\mathbb{E}[\|\nabla f(x^{t})\|^{2}]\leq\epsilon$, where the expectation is over the algorithm. 
\end{definition}

We follow part of the work in SCSG. Based on their algorithm settings, we recall that a random variable $N$ has a geometric distribution $N\sim Geom(\gamma)$ if $N$ is supported on the non-negative integrates, which their elementary calculation has been shown as $\mathbb{E}_{N\sim Geom}(\gamma)=\dfrac{\gamma}{1-\gamma}$. For brevity, we also write $\nabla{f}_{\mathcal{I}}(x)=\dfrac{1}{|\mathcal{I}|}\sum_{i\in\mathcal{I}}\nabla{f}_{i}(x)$. Note that calculating $\nabla{f}_{\mathcal{I}}(x)$ incurs $|\mathcal{I}|$ units of computational cost. The minimum IFO complexity to reach an $\epsilon$-$\textit{accurate}$ solution is denoted by $\mathcal{C}_{comp}(\epsilon)$.

To formulate our complexity bound, we define:
\begin{equation}
f^{*}=\inf_{x}f(x) \quad \text{and} \quad \triangle_{f}=f(\tilde{x}_{0})-f^{*}>0,
\label{ff}
\end{equation}
Further, an upper bound on the variance of the stochastic gradients can be defined as: 
\begin{equation}
\mathcal{S}^{*}=\sup_{x}\dfrac{1}{n}\sum_{i=1}^{n}\|\nabla{f}_{i}(x)-\nabla{f}(x)\|^{2}.
\label{ss}
\end{equation}
\section{Variance controlled SVRG with a combined unbiased/biased estimation}
To resolve the first challenge of SG-based optimization, we provide an adjustable schedule of batch size $B<n$, which scales the sample size for optimization. In the second challenge, the gradient update balance between the full batch and stochastic estimators is fixed. One method~\cite{10.1007/978-3-030-29911-8_26} balanced the gradient of SVRG in terms of the stochastic element and its variance to allow the algorithm to choose appropriate behaviors of gradient from stochastic, through reduced variance, to batch gradient descent by introducing a hyper-parameter $\lambda$. Based on this method, we focus on the $\lambda$ for the subset of full gradients. Towards the last challenge associated with the trade-off between biased and unbiased estimators, we analyze the nature of biased and unbiased estimators in different stages of the non-convex optimization and propose a method that combines the benefits of both biased and unbiased estimator to achieve a fast convergence rate. Firstly, we show a generic form of the batched SVRG in Alg~\ref{a61}, which is proposed by~\cite{NIPS2017_6829}. Compared with the SVRG algorithm, the batched SVRG algorithm has a mini-batch procedure in the inner loop and outputs a random sample instead of an average of the iterates. As seen in the pseudo-code, the batched SVRG method consists of multiple epochs, the batch-size $B_{j}$ is randomly chosen from the whole samples $n$ in $j$-th epoch and work with mini-batch $b_{j}$ to generate the total number of updates for inner $k$-th epoch by a geometric distribution with mean equal to the batch size. Finally it outputs a random sample from $\{\tilde{x}_{j}\}^{T}_{j=1}$. This is a standard way also proposed by~\cite{journals/siamjo/NemirovskiJLS09}, which can save additional overhead by calculating the minimum value of output as $arg\min_{j\leq T\|\nabla{f}(\tilde{x}_{j})\|}$.

\begin{algorithm}[H]	
	\SetKwInOut{Input}{input}\SetKwInOut{Output}{output}
	\Input{Number of epochs $T$, step-size $(\eta_{j})_{j=1}^{T}$, batch size $(B_{j})_{j=1}^{T}$, mini-batch sizes $(b_{j})_{j=1}^{T}$}
	\For {$j=0$ \KwTo $T$ } 
	{Uniformly sample a batch $\mathcal{I}_{j}\subset\{1,...,n\}$ with $|\mathcal{I}_{j}|=B_{j}$\;
	$g_{j}\gets\nabla f_{\mathcal{I}_{j}}(\tilde{x}_{j-1})$\;
        $\tilde{x}_{0}^{(j)}\gets\tilde{x}_{j-1}$\\;
		Generate $\mathcal{N}\sim \text{Geom}(B_{j}/(B_{j}+b_{j}))$\;
		\For {$k=1$ \KwTo $\mathcal{N}_{j}$ } {
			Randomly select $\mathcal{I}_{k-1}\subset\{1,...,n\}$ with $|\tilde{\mathcal{I}}_{k-1}|=b_{j}$\;
			$v_{k-1}^{(j)}=\nabla f_{\tilde{\mathcal{I}}_{k-1}}(x_{k-1}^{(j)})-\nabla{f_{\tilde{\mathcal{I}}_{k-1}}}(x^{(j)}_{0})+g_{j}$; \ \
			$x_{k}^{(j)}= x_{k-1}^{(j)}-\eta_{j}v_{k-1}^{(j)}$\;
		}			
		$\tilde{x}_{j}\gets x_{\mathcal{N}_{j}}^{(j)}$\;
	}
	\Output{Sample $\tilde{x}_{T}^{*}$ from $\{\tilde{x}_{j}\}^{T}_{j=1}$ with $P(\tilde{x}^{*}_{T}=\tilde{x}_{j})\propto\eta_{j}B_{j}/b_{j}$}
	\caption{Batching SVRG}
   \label{a61}
\end{algorithm} 

For the cases of unbiased/biased estimations for the batched SVRG, we provide two great upper bounds on their convergence for their gradients and lower bounds of batch size when their dependency is sample size $n$ the following two sub-sections. Proof details are presented in the appendix.
\subsection{Weighted unbiased estimator on one-epoch analysis}
In the first case, we introduce a hyper-parameter $\lambda$ that is applied in a weighted unbiased version of the batched SVRG and is shown in Alg~\ref{a3}. Since our method based on SVRG, the $\lambda$ should be within the range $0<\lambda<1$ in unbiased and biased cases. Besides, by Lemma~\ref{unlemma1}(see the detail in supplement pages), if $\nabla f_{\tilde{\mathcal{I}}_{k-1}}(x_{k-1}^{(j)})\leq \nabla{f_{\tilde{\mathcal{I}}_{k-1}}}(x^{(j)}_{0})$, then $0<\lambda<\leq\dfrac{1}{2}$. Otherwise, $\dfrac{1}{2}\leq\lambda<1$

\begin{algorithm}[H]
	Replace line number 8 in Alg.~\ref{a61} with the following line: {$v_{k-1}^{(j)}=(1-\lambda)\nabla f_{\tilde{\mathcal{I}}_{k-1}}(x_{k-1}^{(j)})-\lambda\left(\nabla{f_{\tilde{\mathcal{I}}_{k-1}}}(x^{(j)}_{0})-g_{j}\right)$\;}
	\caption{Batching SVRG with weighted unbiased estimator}
		\label{a3}
\end{algorithm}

We now analyse the upper bound of expectation of gradients in a single epoch. Based on the parameter settings by SCSG, we modified two more general formats of schedule including learning rate $\eta_{j}$ and mini-batch size $b_{j}$ to estimate the best schedules in each stage of optimization for both unbiased and biased estimators. Under such settings, we can achieve the upper bound which is shown in~Theorem~\ref{t1u}.
\begin{theorem}
	Let $\eta_{j}L=\gamma(\dfrac{b_{j}}{B_{j}})^{\alpha}$ $(0\leq\alpha\leq 1)$ and $\gamma\geq 0$. Suppose $B_{j}\geq b_{j}\geq B_{j}^{\beta}$ $(0\leq\beta\leq 1)$ for all $j$, then under~Definition~\ref{lsmooth1}, the output $\tilde{x}_{j}$ of Alg~\ref{a3} we have
	\begin{equation}
	\begin{aligned}
	&\mathbb{E}\|\nabla{f}(\tilde{x}_{j})\|^{2}\leq\dfrac{(\dfrac{2L}{\gamma})\triangle_{f}}{{\theta\sum_{j=1}^{T}b_{j}^{\alpha-1}B_{j}^{1-\alpha}}}+\dfrac{2\lambda^{4}I(B_{j}<n)\mathcal{S}^{*}}{\theta B_{j}^{1-2\alpha}},
	\end{aligned}
	\label{unbiased-upperbound}
	\end{equation}
	where $I(B_{j}<n)\geq\dfrac{n-B_{j}}{(n-1)B_{j}}$, $0<\lambda<1$ and $\theta=2(1-\lambda)-(2\gamma B_{j}^{\alpha\beta-\alpha}+2B_{j}^{\beta-1})(1-\lambda)^{2}-1.16(1-\lambda)^{2}$ is positive when $B_{j}\geq 3$ and $0\leq\gamma\leq\tfrac{13}{50}$.
\label{t1u}
\end{theorem}

\subsection{Biased estimator on one-epoch analysis}
In this sub-section we theoretically analyze the performance of the biased estimator, which is shown in Alg~\ref{a6.2}.
\begin{algorithm}[H]
	{Replace the line number 8 in Alg~\ref{a61} with the following line:}
	$v_{k-1}^{(j)}=(1-\lambda)\left(\nabla f_{\tilde{\mathcal{I}}_{k-1}}(x_{k-1}^{(j)})-\nabla{f_{\tilde{\mathcal{I}}_{k-1}}}(x^{(j)}_{0})\right)+\lambda g_{j}$\;
	\caption{Batching SVRG with biased estimator}
	\label{a6.2}
\end{algorithm}
I'm a apple 
Applying the same schedule of $\eta_{j}$ and $b_{j}$ that are used in the unbiased case, we can achieve the result for this case, which is shown in~Theorem~\ref{t1}.
\begin{theorem}
	let $\eta_{j}L=\gamma(\dfrac{b_{j}}{B_{j}})^{\alpha}$ $(0\leq\alpha\leq 1)$ and $0\leq\gamma\leq \dfrac{1}{3}$. Suppose $B_{j}\geq b_{j}\geq B_{j}^{\beta}$ $(0\leq\beta\leq 1)$ for all $j$, then under~Definition~\ref{lsmooth1}, the output $\tilde{x}_{j}$ of Alg~\ref{a6.2} we have,
	\begin{equation}
	\begin{aligned}
	&\mathbb{E}\|\nabla{f}(\tilde{x}_{j})\|^{2}\leq\dfrac{(\dfrac{2L}{\gamma})\triangle_{f}}{{\Theta\sum_{j=1}^{T}b_{j}^{\alpha-1}B_{j}^{1-\alpha}}}+\dfrac{2(1-\lambda)^{2}I(B_{j}<n)\mathcal{S}^{*}}{\Theta B_{j}^{1-2\alpha}},
	\end{aligned}
	\label{biased-upperbound}
	\end{equation}
	where $I(B_{j}<n)\geq\dfrac{n-B_{j}}{(n-1)B_{j}}$, $0<\lambda<1$ and $\Theta=2(1-\lambda)-(2\gamma B_{j}^{\alpha\beta-\alpha}+2B_{j}^{\beta-1}-4LB_{j}^{2\alpha-2})(1-\lambda)^{2}-1.16(1-\lambda)^{2}$. 
\label{t1}
\end{theorem}


\subsection{Convergence analysis on all-epoch}
Over all epochs $T$, the output $\tilde{x}_{T}^{*}$ that is randomly selected from $(\tilde{x}_{j})_{j=1}^{T}$ should be non-convex and $L$-smooth. 
When $1\leq B_{j}=B\leq n$ and $1\leq b_{j}=B_{j}^{\beta}\leq B_{j}$, we can achieve the computational complexity of output from~Theorem~\ref{t1u} and \ref{t1} that is given as
\begin{equation}
\mathbb{E}\|\nabla{f}
(\tilde{x}_{T}^{*})\|^{2}=\mathcal{O}\left(\dfrac{L\triangle_{f}}{TB_{j}^{1+\alpha\beta-\alpha-\beta}}+\dfrac{\mathcal{S}^{*}}{B_{j}^{1-2\alpha}}\right),
\label{upper-bound}
\end{equation}
which covers two extreme cases of complexity bounds since the batch-size $B_{j}$ has two different dependencies. 

We start from considering a constant batch/mini-batch size $B_{j}\equiv B$ for some $1<B\leq n$, $b_{j}\equiv B_{j}^{\beta}$ ($0\leq\beta\leq1$).
\begin{enumerate}
\item \textbf{Dependence on $\epsilon$}. If $b_{j}=B_{j}$ when $\beta=1$ and $1<B_{j}\equiv B<n$, the second term of~Eq.\ref{upper-bound} can be made $\mathcal{O}(\epsilon)$ by setting $B_{j}^{1-2\alpha}=B=\mathcal{O}\left(\dfrac{\mathcal{S}^{*}}{\epsilon}\right)$, where incurs $\alpha=0$. Under such a setting that $B_{j}$ depends on $\epsilon$, $T(\epsilon)=\left(\tfrac{L\triangle_{f}}{\epsilon}\right)$ resulting in the complexity bound is given as $\mathbb{E}C_{comp}(\epsilon)=\mathcal{O}\left(\tfrac{L\triangle_{f}B}{\epsilon}\right)=\mathcal{O}\left(\tfrac{L\triangle_{f}\mathcal{S}^{*}}{\epsilon^{2}}\right),$
which is same with the rate for SGD as shown in Table~\ref{table6.1}. 
\item \textbf{Dependence on n.} If $b_{j}=1$ when $\beta=0$ and $B_{j}=n$, Eq.\ref{upper-bound} can be further alternative as $\mathbb{E}\|\nabla{f}
(\tilde{x}_{T}^{*})\|^{2}=\mathcal{O}\left(\dfrac{L\triangle_{f}}{Tn^{1-\alpha}}+\dfrac{\mathcal{S}^{*}}{n^{1-2\alpha}}\right)$. When $\alpha\leq\dfrac{1}{2}$, $T(\epsilon)$ can be made as $\mathcal{O}\left(1+\tfrac{L\triangle_{f}}{\epsilon n^{1/2}}\right)$, which yields the complexity bound become as 
\begin{equation}
\mathbb{E}C_{comp}(\epsilon)=\mathcal{O}\left(n+\tfrac{n^{\frac{1}{2}}L\triangle_{f}}{\epsilon}\right).
\label{en}
\end{equation} 
This upper bound of rate can guarantee to be better than SCSG, as shown in Table~\ref{table6.1}. 
\end{enumerate}

However, both of the above settings are two sub-optimal cases since their extreme setting either the parameter mini-batch size $b_{j}$ is too large or batch size $B_{j}$ is too large. We now discuss the best parameter schedules over~Theorem.~\ref{t1u} and \ref{t1}, depending on the above two dependencies. 

\textbf{For the case of batch size $B_{j}$ depending on $\epsilon$}, $B_{j}=\mathcal{O}\left(\tfrac{\mathcal{S}^{*}}{\epsilon}\right)$, $b_{j}\neq 1$, and learning rate $\eta_{j}=\tfrac{\gamma}{L}(\tfrac{1}{B_{j}})^{\alpha(1-\beta)}$ where $0\leq\alpha\leq\frac{1}{2}$. To determine the optimal value of $b_{j}$ in this case, we compared to the extreme case when $b_{j}=1$ and $B_{j}=n$ that the optimal schedule of learning rate $\eta_{j}=\tfrac{\gamma}{L}(\tfrac{1}{B_{j}})^{\frac{2}{3}}$ is provided by~\cite{Reddi:2016:SVR:3045390.3045425,7798553,DBLP:conf/icml/ZhuH16,NIPS2017_6829}. Correspondingly in our general form of learning rate, they specified $\alpha=\tfrac{2}{3}$ and $\beta=0$. Thus, the learning rate $\eta_{j}$ has a range which is shown as $\tfrac{\gamma}{L}\geq\tfrac{\gamma}{L}(\tfrac{1}{B_{j}})^{\frac{2}{3}(1-\beta)}\geq\tfrac{\gamma}{L}(\tfrac{1}{B_{j}})^{\frac{1}{2}}$. As a result, we can estimate the range of $\beta$ as $0\leq \beta\leq1/4$. Consequently, $\beta=1/4$ and $\alpha=0$ are the optimal values in this case. 

After determined the three schedules including $B_{j}$, $\eta_{j}$ and $b_{j}$, we can estimate the optimal value of $\lambda^{*}$. For the first case that $B_{j}=\mathcal{O}\left(\tfrac{\mathcal{S}^{*}}{\epsilon}\right)$, $b_{j}=B_{j}^{\frac{1}{4}}$, $\eta_{j}=\tfrac{1}{3L}$, Eq.\ref{upper-bound} is specified as 
\begin{equation}
\mathbb{E}\|\nabla{f}(\tilde{x}_{T}^{*})\|^{2}=\mathcal{O}\left(\dfrac{L\triangle_{f}}{T}\left(\dfrac{\epsilon}{\mathcal{S}^{*}}\right)^{\frac{3}{4}}+\mathcal{\epsilon}\right).
\label{c1}
\end{equation}
Since in this case batch size depends on $\epsilon$, we more focus on the second term in Eq.~\ref{c1}. As a result, we optimize the second term of  $\mathbb{E}\|\nabla{f}(\tilde{x}_{T}^{*})\|^{2}$ from both~Theorem~\ref{t1u} and \ref{t1} in order to achieve lowest upper bound. After comparison the upper bounds in both Eq.~\ref{unbiased-upperbound} and \ref{biased-upperbound}, we choose the optimal value of $\lambda^{*}=(1/16)(15-\sqrt{97})\approx{0.32}$ with the unbiased estimation case, which can provide the lowest upper bound of gradient resulting faster convergence. 

\textbf{For the case of batch size $B_{j}$ depending on $n$}, we now analyse the lower bound of batch size $B_{j}$ in both unbiased and biased estimations. When applying unbiased estimator, for a single epoch, $j$, we define the weighted unbiased variance as $e_{j}=\lambda\left(\nabla{f}_{\mathcal{I}_{j}}(\tilde{x}_{j-1})-\nabla{f}(\tilde{x}_{j-1})\right)$. Thus, the gradients in Alg~\ref{a3} can be updated within the $j$-th epoch as
$\mathbb{E}_{\tilde{\mathcal{I}}_{k}}v_{k}^{(j)}=(1-\lambda)\nabla{f}(x_{k}^{(j)})+e_{j}$, which reveals the key difference between the batched SVRG and the variance controlled batched SVRG on both unbiased/ biased estimators. Most of the novelty in our analysis lies in dealing with the extra term $e_{j}$. Since we achieve a lower bound of batch-size by bounding the term $e_{j}$, we provide the bound of the term $e_{j}$ as $\mathbb{E}_{\mathcal{I}_{j}}\| e_{j}\|^{2}\leq\lambda^{2}\dfrac{n-B_{j}}{nB_{j}}\mathcal{K}^{2}\leq\lambda^{2}\dfrac{n-B_{j}}{nB_{j}}\frac{n}{\sqrt{n-1}}\mathcal{S}^{*}\leq\sigma\rho^{2j}$, where the first inequation follows~\cite{Sampling111,DBLP:journals/corr/BabanezhadAVSKS15} the variance of the norms of gradients $\mathcal{K}^{2}\geq\frac{1}{n-1}\sum_{i=1}^{n}[\|\nabla{f_{i}}(\tilde{x}_{j-1})\|^{2}-\|\nabla{f}(\tilde{x}_{j-1})\|^{2}]$, the second inequation follows the Samuelson inequality~\cite{NIEZGODA2007574} that $\mathcal{K}^{2}\leq\tfrac{n}{\sqrt{n-1}}\mathcal{S}^{*}$ where $\mathcal{S}^{*}$ is shown in Eq.~\ref{ss}, and in the last inequation, there is an upper bound of variance where $\sigma\geq0$ is a constant for some $\rho<1$. Thus $B_{j}$ can be bounded in following theorem.
	\begin{equation}
	B_{j}\geq\dfrac{n\mathcal{S}^{*}}{\mathcal{S}^{*}+\lambda^{2}n^{\frac{1}{2}}\sigma\rho^{2j}}.
	\label{tu-batchsize}
	\end{equation}

For batch size in biased case, we use the same approach adopted in the unbiased version. For a single epoch, $j$, we define the biased variance as $e_{j}=\lambda\nabla{f}_{\mathcal{I}_{j}}(\tilde{x}_{j-1})-(1-\lambda)\nabla{f}(\tilde{x}_{j-1})$. And we achieve the lower bound of batch-size, which is shown in the following. 
	\begin{equation}
		B_{j}\geq\begin{cases}
    	\dfrac{n\mathcal{S}^{*}}{\mathcal{S}^{*}+(1-\lambda)^{2}n^{\frac{1}{2}}\sigma\rho^{2j}}, & \text{if $0<\lambda<\dfrac{\sqrt{2}}{2}$}.\\
    	\dfrac{n\mathcal{S}^{*}}{\mathcal{S}^{*}+(3\lambda^{2}-2\lambda)^{2}n^{\frac{1}{2}}\sigma\rho^{2j}}, & \text{if $\dfrac{\sqrt{2}}{2}<\lambda<1$}.
        \end{cases}
    \label{t-batchsize}
	\end{equation}


To estimate the optimal value of $\lambda^{*}$ in this case that batch size depending on $n$, we specified lower bound of batch size $B_{j}$ which has two versions of biased and unbiased estimations, $b_{j}=1$ and $\eta_{j}=\dfrac{1}{3L}(\dfrac{1}{B_{j}})^{\frac{1}{2}}$ when optimal value $\alpha=\dfrac{1}{2}$. Thus Eq.\ref{upper-bound} can be specified as 
\begin{equation}
\mathbb{E}\|\nabla{f}
(\tilde{x}_{T}^{*})\|^{2}=\mathcal{O}\left(\dfrac{L\triangle_{f}}{TB_{j}^{\frac{1}{2}}}+\mathcal{S}^{*}\right). 
\label{c2}
\end{equation}
Due to this case that batch size depending on $n$, we more focus on the first term in Eq.~\ref{c2}. Thus we optimise the first term in the upper bound of  $\mathbb{E}\|\nabla{f}(\tilde{x}_{T}^{*})\|^{2}$ in both~Theorem~\ref{t1u} and \ref{t1}. After comparison of upper bounds both in unbiased and biased cases (Eq.~\ref{unbiased-upperbound} and \ref{biased-upperbound}, respectively), we determine $\lambda^{*}=5/8$ with biased estimation that obtain the lowest upper bound. 

Consequently, we can achieve the greater complexity bound of Eq.\ref{upper-bound} for both biased/unbiased estimations via replacing full sample size $n$ by the batched sample size $B_{j}$ in Eq.\ref{en}, which is shown in Eq.\ref{complex}.
\begin{equation}
    \mathbb{E}C_{comp}(\epsilon)=\mathcal{O}\left(B+\tfrac{B^{\frac{1}{2}}L\triangle_{f}}{\epsilon}\right)
    \label{complex}
\end{equation}.
\subsection{Best of two worlds}
We have seen in the previous section that the variance controlled SVRG combines the benefits of both SVRG and SGD. We now show these benefits can be made more pronounced by $\lambda^{*}$ with best combinations between $B_{j}$ and $b_{j}$ in different stages of optimization. We introduce our algorithm {\em VCSG} shown in Alg~\ref{a64}. 
\begin{algorithm}[!htb]
 \SetAlgoLined
	\SetKwInOut{Input}{input}
	\SetKwInOut{Output}{output}
\Input{Same input parameters with Alg~\ref{a61} and
$(B_{j})_{j=1}^{T}=\left\{\dfrac{12\mathcal{S}^{*}}{\epsilon}\wedge\dfrac{n\mathcal{S}^{*}}{\mathcal{S}^{*}+0.14\cdot n^{\frac{1}{2}}\sigma\rho^{2j}}\right\}$ where $\sigma\geq 0, \rho<1$}
	\For {$j=1$ \KwTo $T$ } {
	Uniformly sample a batch $\mathcal{I}_{j}\subset\{1,...,n\}$ with $|\mathcal{I}_{j}|=B_{j}$\;
	$g_{j}\gets\nabla f_{\mathcal{I}_{j}}(\tilde{x}_{j-1})$\;
	$\tilde{x}_{0}^{(j)}\gets\tilde{x}_{j-1}$\;
	$\mathcal{S}^{*}\gets(\nabla f_{\mathcal{I}_{j}}(\tilde{x}_{0}^{(j)})-g_{j})^{2}$\;
	Update $B_{j}$\;
	Generate $\mathcal{N}_{j}\sim \text{Geom}(B_{j}/(B_{j}+b_{j}))$\;
			\For {$k=1$\KwTo $\mathcal{N}_{j}$ } {
			\eIf {$B_{j}=\mathcal{S}^{*}/\epsilon$}
			{$b_{j}=B_{j}^{\frac{1}{4}}$; $\eta_{j}=\dfrac{1}{3L}$\;
				Randomly select $\mathcal{I}_{k-1}\subset\{1,...,n\}$ with $|\tilde{\mathcal{I}}_{k-1}|=b_{j}$\;
				\eIf{$\nabla f_{\tilde{\mathcal{I}}_{k-1}}(x_{k-1}^{(j)})<\nabla{f_{\tilde{\mathcal{I}}_{k-1}}}(x^{(j)}_{0})$}
				{$v_{k-1}^{(j)}=(1-\dfrac{1}{16}(15-\sqrt{97}))\cdot\nabla f_{\tilde{\mathcal{I}}_{k-1}}(x_{k-1}^{(j)})
				-\dfrac{1}{16}(15-\sqrt{97})\cdot(\nabla{f_{\tilde{\mathcal{I}}_{k-1}}}(x^{(j)}_{0})-g_{j})$\;}
				{$v_{k-1}^{(j)}=\dfrac{1}{2}\cdot\left(\nabla f_{\tilde{\mathcal{I}}_{k-1}}(x_{k-1}^{(j)})
				-\nabla{f_{\tilde{\mathcal{I}}_{k-1}}}(x^{(j)}_{0})+g_{j}\right)$\;}}
			{$b_{j}=1$; $\eta_{j}=\dfrac{1}{3L}(\dfrac{1}{B_{j}})^{\frac{1}{2}}$\;
				Randomly select $\mathcal{I}_{k-1}\subset\{1,...,n\}$ with $|\tilde{\mathcal{I}}_{k-1}|=b_{j}$\;
				$v_{k-1}^{(j)}=\dfrac{3}{8}\cdot(\nabla f_{\tilde{\mathcal{I}}_{k-1}}(x_{k-1}^{(j)})-\nabla{f_{\tilde{\mathcal{I}}_{k-1}}}(x^{(j)}_{0}))+\dfrac{5}{8}\cdot g_{j}$\;}
				$x_{k}^{(j)}= x_{k-1}^{(j)}-\eta_{j}v_{k-1}^{(j)}$\;}
			$\tilde{x}_{j}\gets x_{\mathcal{N}_{j}}^{(j)}$\;
    }
\Output{Sample $\tilde{x}_{T}^{*}$ from $(\tilde{x}_{j})^{T}_{j=1}$ with $P(\tilde{x}^{*}_{T}=\tilde{x}_{j})\propto\eta_{j}B_{j}/b_{j}$}
\caption{(Mini-Batch)VCSG}
\label{a64}
\end{algorithm}

Following Alg~\ref{a64}, we can achieve a general result for VCSG in the following theorem. 
\begin{theorem}
Suppose $\gamma\leq\dfrac{1}{3}$. Let $B_{j}=\min\left\{\tfrac{\mathcal{S}^{*}}{\epsilon}, \tfrac{n\mathcal{S}^{*}}{\mathcal{S}^{*}+0.14\cdot n^{\frac{1}{2}}\sigma\rho^{2j}}\right\}$, under Definition~\ref{lsmooth1} the output $\tilde{x}^{*}_{T}$ in Alg~\ref{a64} satisfies one of two bounds.
\begin{itemize}
    \item if $B_{j}=\dfrac{\mathcal{S}^{*}}{\epsilon}$,$b_{j}=B_{j}^{\frac{1}{4}}$, $\eta_{j}=\dfrac{\gamma}{L}$ and $\lambda^{*}=(\dfrac{1}{16})(15-\sqrt{97})$ with an unbiased estimator,
\begin{equation*}
\mathbb{E}\|\nabla{f}(\tilde{x}_{T}^{*})\|^{2}
\leq\dfrac{\frac{15.7L}{\gamma}\triangle_{f}}{\sum_{j=1}^{T}B_{j}^{\frac{3}{4}}}+\dfrac{0.17(I(B_{j}<n)\mathcal{S}^{*}}{B_{j}},
\end{equation*}
where $I(B_{j}<n)\geq\dfrac{n-B_{j}}{(n-1)B_{j}}$.
\item if $B_{j}=\dfrac{n\mathcal{S}^{*}}{\mathcal{S}^{*}+0.14\cdot n^{\frac{1}{2}}\sigma\rho^{2j}}$,  $b_{j}=1$, $\eta_{j}=\dfrac{\gamma}{L}(\dfrac{1}{B_{j}})^{\frac{1}{2}}$ and $\lambda^{*}=\dfrac{5}{8}$ with a biased estimator,
    \begin{equation*}
\mathbb{E}\|\nabla{f}(\tilde{x}_{T}^{*})\|^{2}
<\dfrac{\dfrac{5.3L}{\gamma}\triangle_{f}}{\sum_{j=1}^{T}B_{j}^{\frac{1}{2}}}+0.75\mathcal{S}^{*}.
\end{equation*}
\end{itemize}
\label{vcsg}
\end{theorem}

Now we discuss how parameters, including $\lambda$, step-size, batch-size, and mini-batch size, work together to control the variance of gradients from stochastic to batch and balance the trade-off between bias/unbiased estimation in batched optimization. Firstly, in very early iterations $B_{j}$ might choose its first term due to the low variance. In this condition, the small $\lambda$ with relatively large learning rate may help gradients being more stochastic to search more region of problem space, and may help points escape from local minima.  During the increasing variance, the first term of $B_{j}$ would be increased as well, resulting $B_{j}$ will choose its second term and goes to the second case. In the second case, both relatively large $\lambda$, small learning rate and the biased estimator work together that can reduce variance to fast converge into a small region of space. In case of the variance that is reduced too small in the second case,  $B_{j}$ will turn to be its first term. 

To calculate the computational complexity of VCSG, we bring the schedule of batch size $B_{j}$ into Eq.~\ref{complex}, which is shown in~Corollary~\ref{cor4}.
\begin{figure}[!htb]
	\centering
	{
	
		\includegraphics[width=11cm]{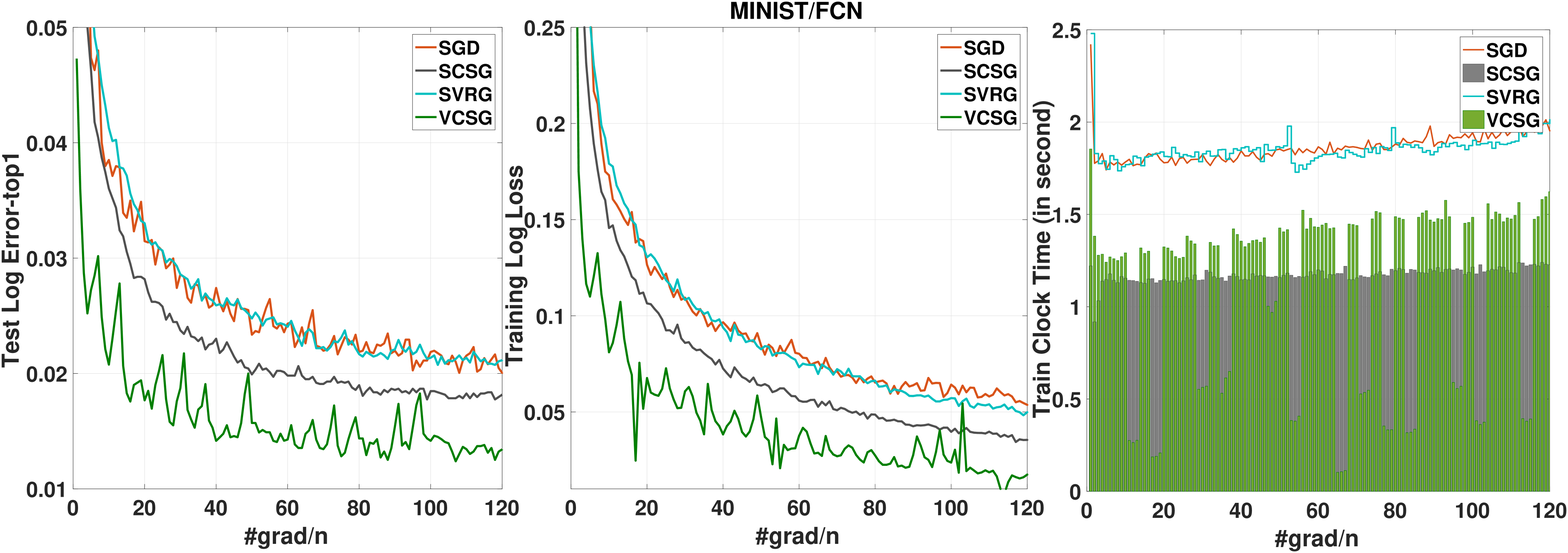}
	}	
	{	
		\includegraphics[width=11cm]{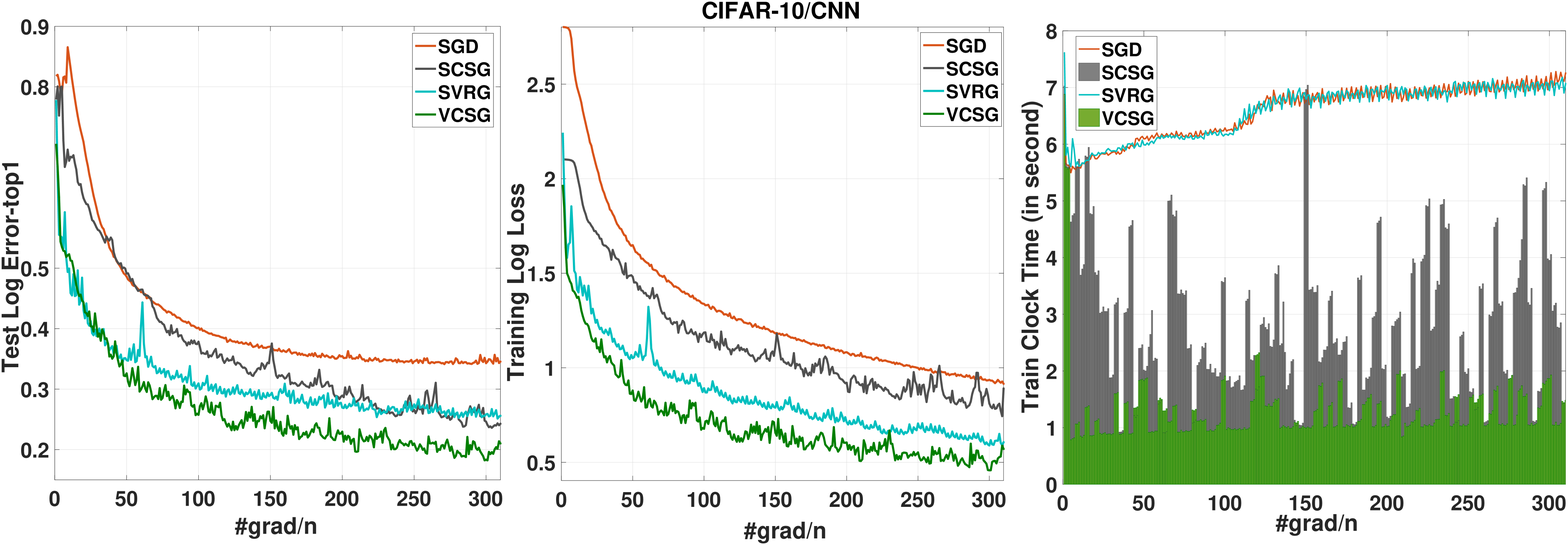}
	}
	{

		\includegraphics[width=11cm]{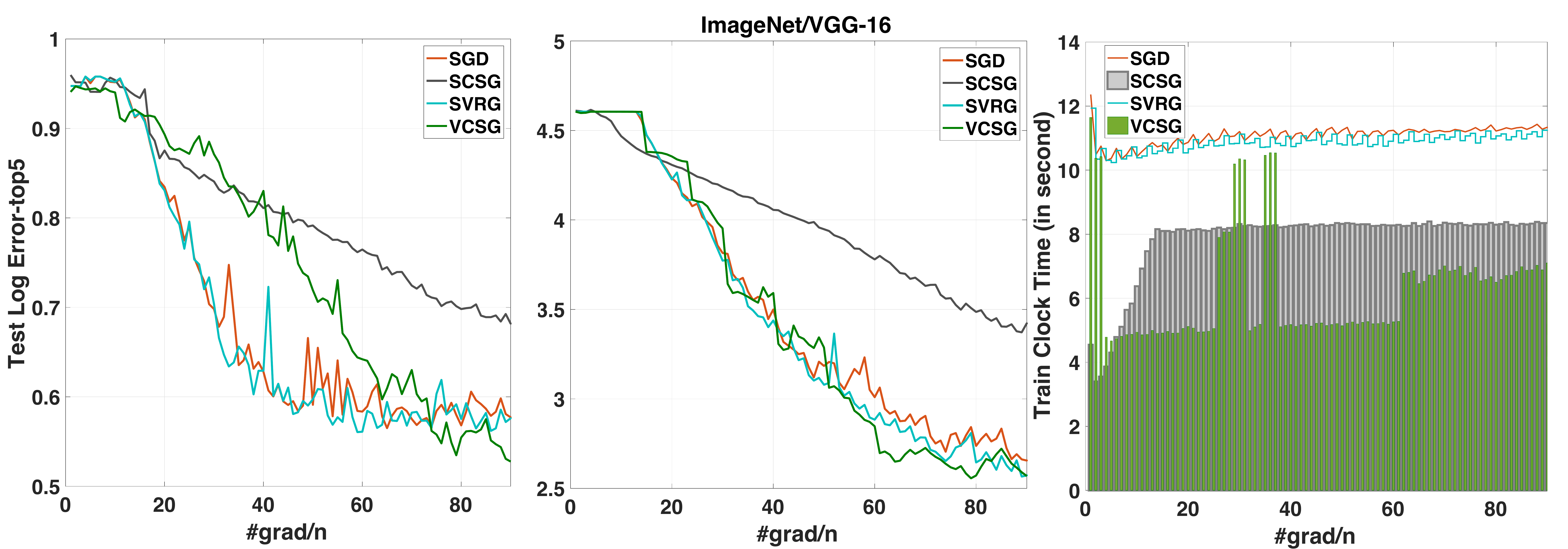}
	}	 	
	\caption{Comparison of rates of convergence in four approaches, including SGD, SVRG, SCSG, and VCSG via test error, training loss and time consumption. Comparatively, we can see that VCSG can converge fastest during all iterations on MINIST and CIFAR-10 data sets. Even though VCSG on the ImageNet data set is slightly slower converging than the other three methods in the beginning, it can significantly decrease after several epochs when the batch-size becomes stable.}
\label{grad}
\end{figure}
\begin{corollary}
	Under parameters setting in~Theorem~\ref{vcsg}, $B_{j}\equiv{B}=\{\tfrac{\mathcal{S}^{*}}{\epsilon}\wedge\tfrac{n\mathcal{S}^{*}}{\mathcal{S}^{*}+0.14\cdot n^{(1/2)}\sigma\rho^{2j}}\}$ then it holds that
	\begin{equation*}
	\begin{aligned}
	\mathbb{E}_{comp}(\epsilon)&=\mathcal{O}\left(B+\dfrac{L\triangle_{f}}{\epsilon}\cdot B^{\frac{1}{2}}\right).
	\end{aligned}
	\end{equation*}
	$B=\{\tfrac{1}{\epsilon}\wedge n^{\frac{1}{2}}\}$ since assume that $L\triangle_{f}, \mathcal{S}^{*}, \sigma\rho^{2j}=\mathcal{O}(1)$. Thus, the above bound can be simplified to 
\begin{equation*}
	\begin{aligned}
	\mathbb{E}_{comp}(\epsilon)&=\mathcal{O}\left((\dfrac{1}{\epsilon}\wedge n^{\frac{1}{2}})+\dfrac{1}{\epsilon}\cdot(\dfrac{1}{\epsilon}\wedge n^{\frac{1}{2}})^{\frac{1}{2}}\right)=\mathcal{O}\left(\dfrac{1}{\epsilon^{\frac{3}{2}}}\wedge \dfrac{n^{\frac{1}{4}}}{\epsilon}\right).
	\end{aligned}
\end{equation*}
\label{cor4}
\end{corollary}

\begin{figure*}[!htb]
	\centering
	\includegraphics[width=11cm]{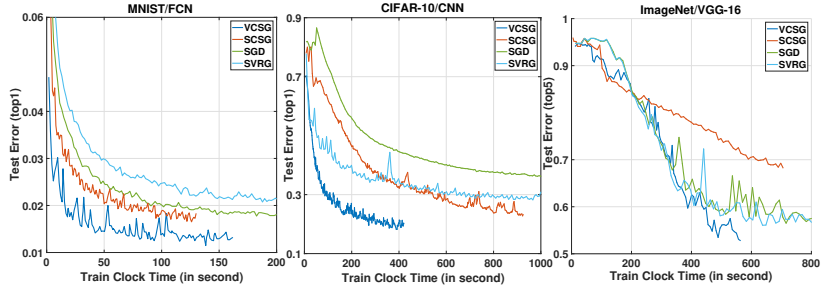}
	\caption{Visualization of test error of four approaches, including SGD, SVRG and SCSG and VCSG against time consumption.}
	\label{timer}
\end{figure*}
\section{Application}
To experimentally verify our theoretical results and insights, we evaluate VCSG compared with SVRG, SGD, and SCSG on three common DL topologies, including LeNet (LeNet-300-100 which has two fully connected layers as hidden layers with 300 and 100 neurons respectively, and LeNet-5 which has two convolutional layers and two fully connected layers) and VGG-16~\cite{journals/corr/SimonyanZ14a} using three datasets including MNIST, CIFAR-10 and tiny ImageNet. Tiny ImageNet contains 200 classes for training each with 500 images and the test set contains 10,000 images. Each image is re-sized to $64\times64$ pixels~\cite{ILSVRC15}.We initialize $B_{j}=B_{0}=n$, correspondingly $b_{j}=b_{0}=n^{\frac{1}{4}}$, $\eta_{0}=1/ (3Ln^{\frac{1}{2}})$, and $\lambda=\tfrac{5}{8}$ via using biased estimator in the first epoch. Meanwhile, we choose a scaled SGD as our baseline by multiplying $0.5$ with stochastic gradients, and applied decayed learning rate $\eta_{j}=\eta_{0}/(j)$ on SGD. For SVRG, we set-up $\lambda=0.5$ with fixed learning rate $\eta_{j}=1/ (3Ln^{\frac{1}{2}})$ in Alg~\ref{a61}. The reason we choose SCSG is that our algorithm is inspired from SCSG which is a leading batched SVRG.

Fig.\ref{grad} compares the performance of four methods, including SGD, SVRG, SCSG, and VCSG, via test log error, training log loss, and training time usage. It has two baselines in all sub-figures, including the performance of SVRG and SGD. The performance of SCSG test error and training loss is smaller than SGD on MNIST and CIFAR-10 data sets, consistent with the experimental results shown in~\cite{NIPS2017_6829}. However, in the ImageNet data set, which is a relatively larger scale application than the previous two data sets, the performance of SCSG becomes worse than SVRG and SGD, which showed weak robustness in our experiments. By contrast, VCSG shown as the green colour in all three datasets, has the lowest test error and training loss among all methods. In the ImageNet data set, both the test error of VCSG is initially higher than SVRG and SGD, but VCSG can reduce the test error and loss dramatically after around 75 epochs. One possible explanation is that the algorithm changes the batch size to the first term resulting in an escape from a local minima by increasing the variance to find a better solution. The right-hand column of~Fig.\ref{grad} presents the time usage, and it can be seen that SVRG and SGD are similar, having higher training time than the other two methods in all three data sets. 

In~Fig.\ref{timer}, we use a more visualized format to show the time usage in~Fig.\ref{grad}. We can see in three sub-figures VCSG can achieve the lowest test error over a shorter time. To achieve the 0.025 top-1 test error in the MNIST data set, VCSG only takes 16 seconds around $2\times$ faster than SCSG, 3$\times$ faster than SVRG, and $4\times$ faster than SGD. In CIFAR-10 to achieve 0.3 top-1 test error, VCSG is around $6\times$ faster than SVRG, $4\times$ faster than SCSG and $13\times$ faster than SGD. In the ImageNet data set, to achieve 0.55 top-5 test error, VCSG can be faster than other methods by up to $5\times$. 

\section{Discussion}
In this paper, we proposed a VR-based optimization $VCSG$ for non-convex problems. We theoretically determined that a hyper-parameter $\lambda$ in each iteration can control the reduced variance of SVRG and balance the trade-off between a biased and an unbiased estimator. Meanwhile, an adjustable batch bounded by controlled reduced variance can work with $\lambda$, step size, and mini-batch to choose an appropriate estimator to converge faster to a stationary point on non-convex problems. Moreover, to verify our theoretical results, our experiments use three datasets on three DL models to present the performance of VCSG via test error/loss and elapsed time and compare these with other leading results. Both theoretical and experimental results show that VCSG can efficiently accelerate convergence. We believe that our algorithm is worthy of further study for non-convex optimization, particularly in deep neural networks training in large-scale applications.
\appendix
\section{Technique lemmas}
The first two lemmas we will use in our theorems are from Lemma A.1 and Lemma A.2 in~\cite{NIPS2017_6829}.  
\begin{lemma}
	Let $x_{1},...,x_{M}\in\mathbb{R}^{d}$ be an arbitrary population of N vectors with 
	\begin{equation*}
	\sum_{j=1}^{M} x_{j}=0.
	\end{equation*}
	Further let $\mathcal{J}$ be a uniform random subset of $\{1,... M\}$ with size m. Then 
	\begin{equation*}
	\begin{aligned}
	\mathbb{E}\|\frac{1}{m}\sum_{j\in\mathcal{J}}x_{j}\|^{2}&=\dfrac{M-m}{(M-1)m}\cdot\frac{1}{M}\sum_{j=1}^{M}\| x_{j}\|^{2}\leq\dfrac{I(m<M)}{m}\cdot\frac{1}{M}\sum_{j=1}^{M}\| x_{j}\|^{2}.
	\end{aligned}
	\end{equation*}
	\label{lemma1}
\end{lemma}

The geometric random variable $N_{j}$ has the key properties below.
\begin{lemma}
	Let $N\sim Geom(\gamma)$ for some $B>0$. Then for any sequence $D_{0},\ D_{1},...,\ D_{N}$ with $\mathbb{E}|D_{N}|<\infty$,
	\begin{equation*}
	\mathbb{E}(D_{N}-D_{N+1})=(\frac{1}{\gamma}-1)(D_{0}-\mathbb{E}D_{N}).
	\end{equation*}
	\label{lemma2}
\end{lemma}

The below lemma will be useful to proof our lemmas. 
\begin{lemma}
	Let $x$ and $y$ $\in\mathbb{R}^{d}$, and a hyper-parameter $0<\lambda<1$. Then, 
	\begin{equation}
	||(1-\lambda)x-\lambda y||^{2}\leq (1-\lambda)^{2}||x-y||^{2},
	\label{lemma-new-eq}
	\end{equation}
	which should satisfied the below conditions:
	\[
    \begin{cases} 	\lambda =(0,\dfrac{1}{2}] & \text{if} \ \ x \leq y \\ 	\lambda =[\dfrac{1}{2},1) & \text{else if} \ \ x \geq y  \end{cases}
    \]
    \begin{proof}
    In the left side of Eq.\ref{lemma-new-eq},
    \begin{equation}
    \begin{aligned}
     ||(1-\lambda)x-\lambda y||^{2}&=(1-\lambda)^{2}||x-\dfrac{\lambda}{1-\lambda} y||^{2}\\
     &=(1-\lambda)^{2}[x^{2}-2(\dfrac{\lambda}{1-\lambda})xy+(\dfrac{\lambda}{1-\lambda})^{2}y^{2})]
    \end{aligned}
    \label{e18}
    \end{equation}
    Let the last line of Eq.~\ref{e18} minus $(1-\lambda)^{2}||x-y||^{2}$, we have
    \begin{equation}
    \begin{aligned}
    &(1-\lambda)^{2}[x^{2}-2(\dfrac{\lambda}{1-\lambda})xy+(\dfrac{\lambda}{1-\lambda})^{2}y^{2}]-(1-\lambda)^{2}||x-y||^{2}\\
    &=(1-\lambda)^{2}[x^{2}-2(\dfrac{\lambda}{1-\lambda})xy+(\dfrac{\lambda}{1-\lambda})^{2}y^{2}-x^{2}+2xy-y^{2}]\\
    &=(1-\lambda)^{2}\left([(\dfrac{\lambda}{1-\lambda})^{2}-1]y^{2}-2[(\dfrac{\lambda}{1-\lambda})-1]xy\right)\\
    &=(1-\lambda)^{2}(\dfrac{\lambda}{1-\lambda}-1)\left([(\dfrac{\lambda}{1-\lambda})^{2}+1]y^{2}-2xy\right).
    \end{aligned}
    \label{eq19}
    \end{equation}
    Eq.~\ref{lemma-new-eq} can be hold when the above equation less than $0$. The first term in the last line of Eq.~\ref{eq19} $(1-\lambda)^{2}$ is positive. For the second and third term, it has two cases. If the first case that the second term $(\dfrac{\lambda}{1-\lambda}-1)\geq 0$ as $\dfrac{1}{2}\leq \lambda<1$, Eq.~\ref{lemma-new-eq} can be hold when third term $[(\dfrac{\lambda}{1-\lambda})^{2}+1]y^{2}-2xy\leq 0$. Thus $x\geq y$ in the first case. Else if the second case that the term $(\dfrac{\lambda}{1-\lambda}-1)\leq 0$ as $0< \lambda\leq\dfrac{1}{2}$, Eq.~\ref{lemma-new-eq} can be hold when third term $[(\dfrac{\lambda}{1-\lambda})^{2}+1]y^{2}-2xy\geq 0$. Thus $x\leq y$ in the second case.
    \end{proof}
	\label{lemma-new}
\end{lemma}

\section{One-Epoch Analysis}
\subsection{Unbiased Estimator Version}
Our algorithm is based on the SVRG method, thus the hyper-parameter $\lambda$ should be within the range as $0<\lambda<1$ in both unbiased and biased cases. We  provide all useful lemmas we will applied in our proof of theorems at first, and provide a proof sketch for  guidance. We start by bounding the gradient $\mathbb{E}_{\mathcal{\tilde{I}}_{k}}\parallel v_{k}^{(j)}\parallel^{2}$ in~Lemma~\ref{unlemma1} and the variance $\mathbb{E}_{\mathcal{I}_{j}}\parallel e_{j}\parallel^{2}$ in~Lemma~\ref{unlemma2}.

\begin{lemma}\label{unlemma1}
	Under~Definition~\ref{lsmooth1}, 
	\begin{equation*}
	\begin{aligned}
	&\mathbb{E}_{\mathcal{\tilde{I}}_{k}}\parallel v_{k}^{(j)}\parallel^{2}\leq \dfrac{L^{2}}{b_{j}}\parallel x_{k}^{(j)}- x_{0}^{(j)}\parallel +2(1-\lambda)^{2}\parallel\nabla f(x_{k}^{(j)})\parallel^{2}+2\lambda^{2}\parallel e_{j}\parallel^{2}.
	\end{aligned}
	\end{equation*}
	\begin{proof}
		Using the fact that for a random variable Z $\mathbb{E}\parallel Z\parallel^{2}=\mathbb\parallel Z-\mathbb{E}Z\parallel^{2}+\parallel \mathbb{E}Z\parallel^{2}$, we have 
		\begin{equation}
		\begin{aligned}
		\mathbb{E}_{\mathcal{\tilde{I}}_{k}}\parallel v_{k}^{(j)}\parallel^{2}=&\mathbb{E}_{\mathcal{\tilde{I}}_{k}}\parallel v_{k}^{(j)}-\mathbb{E}_{\mathcal{\tilde{I}}_{k}} v_{k}^{(j)}\parallel^{2}+\parallel\mathbb{E}_{\mathcal{\tilde{I}}_{k}} v_{k}^{(j)}\parallel^{2}\\
		=&\mathbb{E}_{\mathcal{\tilde{I}}_{k}}\parallel(1-\lambda)\nabla f_{\mathcal{\tilde{I}}_{k}}(x_{k}^{(j)})-\lambda\nabla f_{\mathcal{\tilde{I}}_{k}}(x_{0}^{(j)})-((1-\lambda)\nabla{f}(x_{k}^{(j)})-\lambda\nabla{f}x_{0}^{(j)})\parallel^{2}\\
		&+\parallel(1-\lambda)\nabla{f}(x_{k}^{(j)})+\lambda e_{j}\parallel^{2}\\
		\leq&\mathbb{E}_{\mathcal{\tilde{I}}_{k}}\parallel(1-\lambda)\nabla{f}_{\mathcal{\tilde{I}}_{k}}(x_{k}^{(j)})-\lambda\nabla{f}_{\mathcal{\tilde{I}}_{k}}(x_{0}^{(j)})-((1-\lambda)\nabla{f}(x_{k}^{(j)})-\lambda\nabla{f}x_{0}^{(j)})\parallel^{2}\\
		&+2\parallel(1-\lambda)\nabla{f}(x_{k}^{(j)})\parallel^{2}+2\parallel\lambda e_{j}\parallel^{2}.
		\end{aligned}
		\end{equation}
		By~Lemma~\ref{lemma1},
		\begin{equation}
		\begin{aligned}
		&\mathbb{E}_{\mathcal{\tilde{I}}_{k}}\parallel(1-\lambda)\nabla{f}_{\mathcal{\tilde{I}}_{k}}(x_{k}^{(j)})-\lambda\nabla{f}_{\mathcal{\tilde{I}}_{k}}(x_{0}^{(j)})-((1-\lambda)\nabla{f}(x_{k}^{(j)})-\lambda\nabla{f}x_{0}^{(j)})\parallel^{2}\\
		&\leq\frac{1}{b_{j}}\cdot\frac{1}{n}\sum_{i=1}^{n}\parallel(1-\lambda)\nabla{f}_{i}(x_{k}^{(j)})-\lambda\nabla{f}_{i}(x_{0}^{(j)})-((1-\lambda)\nabla{f}(x_{k}^{(j)})-\lambda\nabla{f}(x_{0}^{(j)}))\parallel^{2}\\
		&=\frac{1}{b_{j}}\cdot(\frac{1}{n}\sum_{i=1}^{n}\parallel(1-\lambda)\nabla{f}_{i}(x_{k}^{(j)})-\lambda\nabla{f}_{i}(x_{0}^{(j)})\parallel^{2}-\parallel((1-\lambda)\nabla{f}(x_{k}^{(j)})-\lambda\nabla{f}(x_{0}^{(j)}))\parallel^{2})\\
		&\leq\frac{1}{b_{j}}\cdot\frac{1}{n}\sum_{i=1}^{n}\parallel(1-\lambda)\nabla{f}_{i}(x_{k}^{(j)})-\lambda\nabla{f}_{i}(x_{0}^{(j)})\parallel^{2}\\
		&\leq\frac{1}{b_{j}}\cdot\frac{(1-\lambda)^{2}}{n}\sum_{i=1}^{n}\parallel\nabla{f}_{i}(x_{k}^{(j)})-\nabla{f}_{i}(x_{0}^{(j)})\parallel^{2}\\
		&\leq\frac{1}{b_{j}}\cdot L^{2}\parallel x_{k}^{(j)}- x_{0}^{(j)}\parallel^{2}
		\end{aligned}
		\end{equation}
		where the last but one line is used Lemma~\ref{lemma-new} when satisfied the condition that if $\nabla{f}_{i}(x_{k}^{(j)})\leq \nabla{f}_{i}(x_{0}^{(j)}$, $\lambda$ should be in the range $(0,\dfrac{1}{2}]$. Otherwise, $\lambda$ should be in the range $[\dfrac{1}{2},1)$. And the last line is based on~Definition~\ref{lsmooth1}, then the bound of the gradient can be alternatively written as, 
		\begin{equation}
		\begin{aligned}
		&\mathbb{E}_{\mathcal{\tilde{I}}_{k}}\parallel v_{k}^{(j)}\parallel^{2}\leq\frac{L^{2}}{b_{j}}\parallel x_{k}^{(j)}- x_{0}^{(j)}\parallel^{2}+2(1-\lambda)^{2}\parallel\nabla{f}(x_{k}^{(j)})\parallel^{2}+2\lambda^{2}\parallel e_{j}\parallel^{2}.
		\end{aligned}
		\end{equation}
	\end{proof}
\end{lemma}

\begin{lemma}
	\begin{equation*}
	\mathbb{E}_{\mathcal{I}_{j}}\parallel e_{j}\parallel^{2}\leq\lambda^{2}\dfrac{I(B_{j}<n)}{B_{j}}\cdot\mathcal{S^{*}}.
	\end{equation*}
	\begin{proof}
		Based on~Lemma~\ref{unlemma1} and the observation that $\tilde{x}_{j-1}$ is independent of $\mathcal{I}_{j}$, the bound of variance $e_{j}$ can be expressed as
		\begin{equation}
		\begin{aligned}
		\mathbb{E}_{\mathcal{I}_{j}}\parallel e_{j}\parallel^{2}&=\dfrac{n-B_{j}}{(n-1)B_{j}}\cdot\frac{\lambda^{2}}{n}\sum_{i=1}^{n}\parallel\nabla{f_{i}}(\tilde{x}_{j-1})-\nabla{f}(\tilde{x}_{j-1})\parallel^{2}\\
		&\leq\lambda^{2}\dfrac{n-B_{j}}{(n-1)B_{j}}\cdot\mathcal{S^{*}}\leq\lambda^{2}\dfrac{I(B_{j}<n)}{B_{j}}\mathcal{S^{*}}
		\end{aligned}
		\label{unlemma3-e1}
		\end{equation}
		where the upper bound of the variance of the stochastic gradients $\mathcal{S^{*}}=\frac{1}{n}\sum_{i=1}^{n}\parallel\nabla{f_{i}}(\tilde{x}_{j-1})-\nabla{f}(\tilde{x}_{j-1})\parallel^{2}$.
	\end{proof}
	\label{unlemma2}
\end{lemma}
\begin{lemma}
	\label{unlemma6} 
	Suppose $\eta_{j}L<1$, then under~Definition~\ref{lsmooth1}, 
	\begin{equation*}
	\begin{aligned}
	&(1-\lambda)\eta_{j}(1-(1-\lambda)L\eta_{j})B_{j}\mathbb{E}\parallel\nabla{f}(\tilde{x}_{j})\parallel^{2}+\lambda\eta_{j}B_{j}\mathbb{E}<e_{j},\nabla{f}(\tilde{x}_{j})>\\
	&\leq b_{j}\mathbb{E}(f(\tilde{x}_{j-1})-f(\tilde{x}_{j}))+\dfrac{\eta_{j}^{2}B_{j}L^{3}}{2b_{j}}\mathbb{E}\parallel\tilde{x}_{j}-\tilde{x}_{j-1}\parallel^{2}+\lambda^{2}L\eta_{j}^{2}B_{j}\mathbb{E}\parallel e_{j}\parallel^{2}.
	\end{aligned}
	\label{unelemma6}
	\end{equation*}
	where $\mathbb{E}$ denotes the expectation with respect to all randomness. 
	\begin{proof}
		By~Definition~\ref{lsmooth1}, we have
		\begin{equation}
		\begin{aligned}
		&\mathbb{E}_{\tilde{\mathcal{I}}_{k}}[f(x_{k+1}^{(j)})]\leq f(x_{k}^{(j)})-\eta_{j}<\mathbb{E}_{\tilde{\mathcal{I}}_{k}}v_{k},\nabla{f}(x_{k}^{(j)})>+\dfrac{L\eta_{j}^{2}}{2}\mathbb{E}_{\tilde{\mathcal{I}}_{k}}\parallel v_{k}\parallel^{2}\\
		&=f(x_{k}^{(j)})-\eta_{j}<((1-\lambda)\nabla{f}(x_{k}^{(j)})+\lambda e_{j}),\nabla{f}(x)_{k}^{(j)})>+\dfrac{L\eta_{j}^{2}}{2}\mathbb{E}_{\tilde{\mathcal{I}}_{k}}\parallel v_{k}\parallel^{2}\\
		&\leq f(x_{k}^{(j)})-\eta_{j}(1-\lambda)\parallel\nabla{f}(x_{k}^{(j)})\parallel^{2}-\eta_{j}<\lambda e_{j},\nabla{f}(x_{k}^{(j)})>+\dfrac{L^{3}\eta_{j}^{2}}{2b_{j}}\parallel (1-\lambda)x_{k}^{(j)}-\lambda x_{0}^{(j)}\parallel^{2}\\
		&+L\eta_{j}^{2}(1-\lambda)^{2}\parallel\nabla{f}(x_{k}^{(j)})\parallel^{2}+L\eta_{j}^{2}\lambda^{2}\parallel e_{j}\parallel^{2}\\
		&=f(x_{k}^{(j)})-(\eta_{j}(1-\lambda)-L\eta_{j}^{2}(1-\lambda)^{2})\parallel\nabla{f}(x_{k}^{(j)})\parallel^{2}-\lambda\eta_{j}<e_{j},\nabla{f}(x_{k}^{(j)})>\\
		&+\dfrac{L^{3}\eta_{j}^{2}}{2b_{j}}\parallel(1-\lambda)x_{k}^{(j)}-\lambda x_{0}^{(j)}\parallel^{2}+L\eta_{j}^{2}\lambda^{2}\parallel e_{j}\parallel^{2}\\
		&\leq f(x_{k}^{(j)})-(\eta_{j}(1-\lambda)-L\eta_{j}^{2}(1-\lambda)^{2})\parallel\nabla{f}(x_{k}^{(j)})\parallel^{2}-\lambda\eta_{j}<e_{j},\nabla{f}(x_{k}^{(j)})>\\
		&+\dfrac{L^{3}\eta_{j}^{2}}{2b_{j}}\parallel x_{k}^{(j)}- x_{0}^{(j)}\parallel^{2}+L\eta_{j}^{2}\lambda^{2}\parallel e_{j}\parallel^{2}\\
		\end{aligned}
		\label{unle5}
		\end{equation}
		Let $\mathbb{E}_{j}$ denote the expectation $\tilde{\mathcal{I}}_{0}, \tilde{\mathcal{I}}_{1}$,..., given $\tilde{\mathcal{N}}_{j}$ since $\tilde{\mathcal{N}}_{j}$ is independent of them and let k=$\mathcal{N}_{j}$ in Inq.~\ref{unle5}. As $\tilde{\mathcal{I}}_{k+1}, \tilde{\mathcal{I}}_{k+2}$,... are independent of $x_{k}^{(j)}$ and taking the expectation with respect to $\mathcal{N}_{j}$ and using Fubini's theorem, Inq.~\ref{unle5} implies that 
		\begin{equation}
		\begin{aligned}
		&\eta_{j}(1-\lambda)(1-(1-\lambda)L\eta_{j})\mathbb{E}_{\mathcal{N}_{j}}\mathbb{E}_{j}[\parallel\nabla{f}(x_{\mathcal{N}_{j}}^{(j)})\parallel^{2}]+\lambda\eta_{j}\mathbb{E}_{\mathcal{N}_{j}}\mathbb{E}_{j}<e_{j},\nabla{f}(x_{\mathcal{N}_{j}}^{(j)})>\\
		&\leq\mathbb{E}_{\mathcal{N}_{j}}(\mathbb{E}_{j}[f(x_{\mathcal{N}_{j}}^{(j)})]-\mathbb{E}_{j}[f(x_{\mathcal{N}_{j+1}}^{(j)})])+\dfrac{L^{3}\eta_{j}^{2}}{2b_{j}}\mathbb{E}_{\mathcal{N}_{j}}\mathbb{E}_{j}\mathbb{E}[\parallel (1-\lambda)x_{\mathcal{N}_{j}}^{(j)}-\lambda x_{0}^{(j)}\parallel^{2}]+L\lambda^{2}\eta_{j}^{2}\parallel e_{j}\parallel^{2}\\
		&=\dfrac{b_{j}}{B_{j}}(f(x_{0}^{(j)})-\mathbb{E}_{j}\mathbb{E}_{\mathcal{N}_{j}}[f_{\mathcal{N}_{j}}^{(j)}])+\dfrac{L^{3}\eta_{j}^{2}}{2b_{j}}\mathbb{E}_{j}\mathbb{E}_{\mathcal{N}_{j}}[\parallel (1-\lambda)x_{\mathcal{N}_{j}}^{(j)}-\lambda x_{0}^{(j)}\parallel^{2}]+L\lambda^{2}\eta_{j}^{2}\parallel e_{j}\parallel^{2}
		\end{aligned}  
		\label{unle6}
		\end{equation}
		where the last equation in Inq.~\ref{unle6} follows from~Lemma~\ref{lemma2}. The lemma substitutes $x_{\mathcal{N}_{j}}^{(j)}(x_{0}^{j})$ by $\tilde{x}_{j}(\tilde{x}_{j-1})$. 
	\end{proof}
\end{lemma}	

\begin{lemma}
	\label{unlemma7}
	Suppose $\eta_{j}^{2}L^{2}B_{j}<b_{j}^{2}$, then under~Definition~\ref{lsmooth1}, 
	\begin{equation*}
	\begin{aligned}
	&(b_{j}-\dfrac{\eta_{j}^{2}L^{2}B_{j}}{b_{j}})\mathbb{E}[\parallel\tilde{x}_{j}-\tilde{x}_{j-1}\parallel^{2}]+2\lambda\eta_{j}B_{j}\mathbb{E}<e_{j},(\tilde{x}_{j}-\tilde{x}_{j-1})>\\
	&\leq-2\eta_{j}(1-\lambda)B_{j}\mathbb{E}<\nabla{f}(\tilde{x}_{j}),(\tilde{x}_{j}-\tilde{x}_{j-1})>+2(1-\lambda)^{2}\eta_{j}^{2}B_{j}\mathbb{E}[\parallel\nabla{f}(\tilde{x}_{j})\parallel^{2}]+2\lambda^{2}\eta_{j}^{2}B_{j}\mathbb{E}[\parallel e_{j}\parallel^{2}]
	\end{aligned}  
	\label{unelemma7}
	\end{equation*}
	\begin{proof}
		Since $x_{k+1}^{(j)}=x_{k}^{(j)}-\eta_{j}v_{k}^{(j)}$, we have 
		\begin{equation}
		\begin{aligned}
		&\mathbb{E}_{\mathcal{\tilde{I}}_{k}}[\parallel x_{k+1}^{(j)}-x_{0}^{(j)}\parallel^{2}]\\
		&=\parallel x_{k}^{(j)}-x_{0}^{(j)}\parallel^{2}-2\eta_{j}<\mathbb{E}_{\mathcal{\tilde{I}}_{k}}v_{k}^{(j)},(x_{k}^{(j)}-x_{0}^{(j)})>+\eta^{2}_{j}\mathbb{E}_{\mathcal{\tilde{I}}_{k}}\parallel v_{k}^{(j)}\parallel^{2}\\
		&=\parallel x_{k}^{(j)}-x_{0}^{(j)}\parallel^{2}-2(1-\lambda)\eta_{j}<\nabla{f}(x_{k}^{(j)}),(x_{k}^{(j)}-x_{0}^{(j)})>-2\lambda\eta_{j}<e_{j},(x_{k}^{(j)}-x_{0}^{(j)})>+\eta_{j}^{2}\mathbb{E}_{\mathcal{\tilde{I}}_{k}}\parallel v_{k}^{(j)}\parallel^{2}\\
		&\leq(1+\dfrac{\eta_{j}^{2}L^{2}}{b_{j}})\parallel x_{k}^{(j)}-x_{0}^{(j)}\parallel^{2}-2\eta_{j}(1-\lambda)<\nabla{f}(x_{k}^{(j)}), x_{k}^{(j)}-x_{0}^{(j)}>\\
		&-2\lambda\eta_{j}<e_{j},(x_{k}^{(j)}-x_{0}^{(j)})>+2(1-\lambda)^{2}\eta_{j}^{2}\parallel\nabla{f}(x_{k}^{(j)})\parallel^{2}+2\lambda^{2}\eta_{j}^{2}\parallel e_{j}\parallel^{2}.
		\end{aligned}
		\end{equation}
		where the last inequality follows from~Lemma~\ref{unlemma1}. Using the same notation $\mathbb{E}_{j}$ from~Eq~\ref{tu-batchsize} we have
		\begin{equation}
		\begin{aligned}
		&2\eta_{j}(1-\lambda)\mathbb{E}_{j}<\nabla{f}(x_{k}^{(j)}),(x_{k}^{(j)}-x_{0}^{(j)})>+2\lambda\eta_{j}\mathbb{E}_{j}<e_{j},(x_{k}^{(j)}-x_{0}^{(j)})>\\
		&\leq(1+\dfrac{\eta_{j}^{2}L^{2}}{b_{j}})\mathbb{E}_{j}\parallel x_{k}^{(j)}-x_{0}^{(j)}\parallel^{2}-\mathbb{E}_{j}\parallel x_{k+1}^{(j)}-x_{0}^{(j)}\parallel^{2}+2(1-\lambda)^{2}\eta_{j}^{2}\parallel \nabla{f}(x_{k}^{(j)})\parallel^{2}+2\lambda\eta_{j}^{2}\parallel e_{j}\parallel^{2}
		\end{aligned}
		\end{equation}
		Let $k=N_{j}$, and using Fubini's theorem, we have, 
		\begin{equation}
		\begin{aligned}
		&2(1-\lambda)\eta_{j}\mathbb{E}_{N_{j}}\mathbb{E}_{j}<\nabla{f}(x_{N_{j}}^{(j)}),(x_{N_{j}}^{(j)}-x_{0}^{(j)})>+2\lambda\eta_{j}\mathbb{E}_{N_{j}}\mathbb{E}_{j}<e_{j},(x_{N_{j}}^{(j)}-x_{0}^{(j)})>\\
		&\leq(1+\dfrac{\eta_{j}L^{2}}{b_{j}})\mathbb{E}_{N_{j}}\mathbb{E}_{j}\parallel x_{N_{j}}^{(j)}-x_{0}^{(j)}\parallel^{2}-\mathbb{E}_{N_{j}}\mathbb{E}_{j}\parallel x_{N_{j}+1}^{(j)}-x_{0}^{(j)}\parallel^{2}\\
		&+2(1-\lambda)^{2}\eta_{j}^{2}\mathbb{E}_{N_{j}}\parallel\nabla{f}(x_{N_{j}}^{(j)})\parallel^{2}+2\lambda^{2}\eta_{j}^{2}\parallel e_{j}\parallel^{2}\\
		&=(-\dfrac{b_{j}}{B_{j}}+\dfrac{\eta_{j}^{2}L^{2}}{b_{j}})\mathbb{E}_{N_{j}}\mathbb{E}_{j}\parallel x_{N_{j}}^{(j)}-x_{0}^{(j)}\parallel^{2}+2(1-\lambda)^{2}\eta^{2}_{j}\mathbb{E}_{N_{j}}\parallel\nabla{f}(x_{N_{j}}^{(j)})\parallel^{2}+2\lambda^{2}\eta_{j}^{2}\parallel e_{j}\parallel^{2}.
		\end{aligned}
		\end{equation}
		The lemma is then proved by substituting $x_{N_{j}}^{(j)}(x_{0}^{(j)})$ by $\tilde{x}_{j}(\tilde{x}_{j-1})$.
	\end{proof}
\end{lemma}
\begin{lemma}
	\label{unlemma8}
	\begin{equation*}
	\begin{aligned}
	&b_{j}\mathbb{E}<e_{j},(\tilde{x}_{j}-\tilde{x}_{j-1})>=-\eta_{j}(1-\lambda)B_{j}\mathbb{E}<e_{j},\nabla{f}(\tilde{x}_{j})>-\lambda^{2}\eta_{j}B_{j}\mathbb{E}\parallel e_{j}\parallel^{2}
	\end{aligned}
	\end{equation*}
	\begin{proof}
		Let $M_{k}^{(j)}=<e_{j},(x_{k}^{(j)}-x_{0}^{(j)})>$, then we have
		\begin{equation}
		\mathbb{E}_{N_{j}}<e_{j},(\tilde{x}_{j}-\tilde{x}_{j-1})>=\mathbb{E}_{N_{j}}M_{N_{j}}^{(j)}.
		\end{equation}
		Since $N_{j}$ is independent of $(x_{0}^{(j)}, e_{j})$, it has 
		\begin{equation}
		\mathbb{E}<e_{j},(\tilde{x}_{j}-\tilde{x}_{j-1})>=\mathbb{E}M_{N_{j}}^{(j)}.
		\end{equation}
		Also $M_{0}^{(j)}=0$, then we have 
		\begin{equation}
		\begin{aligned}
		&\mathbb{E}_{\mathcal{\tilde{I}}_{k}}(M_{k+1}^{(j)}-M_{k}^{(j)})\\
		&=\mathbb{E}_{\mathcal{\tilde{I}}_{k}}<e_{j},(x_{k+1}^{(j)}-x_{k}^{(j)})>\\
		&=-\eta_{j}<e_{j},\mathbb{E}_{\mathcal{\tilde{I}}_{k}}[v_{k}^{(j)}]>.
		\end{aligned}
		\end{equation}
		Using the same notation $\mathbb{E}_{j}$ in~Lemma~\ref{unlemma6} and~Lemma~\ref{unlemma7}, we have
		\begin{equation}
		\begin{aligned}
		&\mathbb{E}_{j}(M_{k+1}^{(j)}-M_{k}^{(j)})=-\eta_{j}(1-\lambda)<e_{j},\mathbb{E}_{j}\nabla{f}(x_{k}^{(j)})>-\lambda^{2}\eta_{j}\parallel e_{j}\parallel^{2}.
		\end{aligned}
		\label{unelemma8}
		\end{equation}
		Let $k=N_{j}$ in~Eq.\ref{unelemma8}. Using Fubini's theorem and~Lemma~\ref{unlemma2}, we have,
		\begin{equation}
		\begin{aligned}
		&\dfrac{b_{j}}{B_{j}}\mathbb{E}_{N_{j}}M_{N_{j}}^{(j)}=-\eta_{j}(1-\lambda)<e_{j},\mathbb{E}_{N_{j}}\mathbb{E}_{j}\nabla{f}(x_{k}^{(j)})>-\eta_{j}\parallel e_{j}\parallel^{2}.
		\end{aligned}
		\end{equation}
		The lemma is then proved by substituting $x_{N_{j}}^{(j)}(x_{0}^{(j)})$ by $\tilde{x}_{j}(\tilde{x}_{j-1})$.
	\end{proof}
\end{lemma}	

\section*{Proof of~Theorem~\ref{t1u}}
\begin{theorem-non}
		Let $\eta_{j}L=\gamma(\dfrac{b_{j}}{B_{j}})^{\alpha}$ $(0\leq\alpha\leq 1)$ and $\gamma\geq 0$. Suppose $B_{j}\geq b_{j}\geq B_{j}^{\beta}$ $(0\leq\beta\leq 1)$ for all $j$, then under~Definition~\ref{lsmooth1}, the output $\tilde{x}_{j}$ of Alg~\ref{a3} we have
	\begin{equation*}
	\begin{aligned}
	&\mathbb{E}\|\nabla{f}(\tilde{x}_{j})\|^{2}\leq\dfrac{(\dfrac{2L}{\gamma})\triangle_{f}}{{\theta\sum_{j=1}^{T}b_{j}^{\alpha-1}B_{j}^{1-\alpha}}}+\dfrac{2\lambda^{4}I(B_{j}<n))\mathcal{S}^{*}}{\theta B_{j}^{1-2\alpha}},
	\end{aligned}
	\end{equation*}
	where $0<\lambda<1$ and $\theta=2(1-\lambda)-(2\gamma B_{j}^{\alpha\beta-\alpha}+2B_{j}^{\beta-1})(1-\lambda)^{2}-1.16(1-\lambda)^{2}$ is positive when $B_{j}\geq 3$ and $0\leq\gamma\leq\dfrac{13}{50}$.
	\begin{tcolorbox}
		\textit{Proof Sketch}: Combine two equations in Lemma~\ref{unlemma6} and Lemma~\ref{unlemma7}, we can achieve a upper bound of unbiased version gradient in single epoch. And further use Lemma~\ref{unlemma2}, the final result of~Theorem~\ref{t1u} can be achieved. 
	\end{tcolorbox}
	\begin{proof}
		Multiplying~Eq.\ref{unelemma6} by 2 and~Eq.\ref{unelemma7} by $\dfrac{b_{j}}{\eta_{j}B_{j}}$ and summing them, then we have,
		\begin{equation}
		\begin{aligned}
		&2\eta_{j}B_{j}(1-\lambda)(1-(1-\lambda)L\eta_{j}-\dfrac{(1-\lambda)b_{j}}{B_{j}})\mathbb{E}\parallel\nabla{f}(\tilde{x}_{j})\parallel^{2}+\dfrac{b_{j}^{3}-\eta_{j}^{2}L^{2}b_{j}B_{j}-\eta_{j}^{3}L^{3}B_{j}^{2}}{b_{j}\eta_{j}B_{j}}\mathbb{E}\parallel\tilde{x}_{j}-\tilde{x}_{j-1}\parallel^{2}\\
		&+2\lambda\eta_{j}B_{j}\mathbb{E}<e_{j},\nabla{f}(\tilde{x}_{j})>+2\lambda b_{j}\mathbb{E}<e_{j},(\tilde{x}_{j}-\tilde{x}_{j-1})>\\
		&=2\eta_{j}B_{j}(1-\lambda)(1-(1-\lambda)L\eta_{j}-\dfrac{(1-\lambda)b_{j}}{B_{j}})\mathbb{E}\parallel\nabla{f}(\tilde{x}_{j})\parallel^{2}\\
		&+\dfrac{b_{j}^{3}-(1-\lambda)^{2}\eta_{j}^{2}L^{2}b_{j}B_{j}-(1-\lambda)^{2}\eta_{j}^{3}L^{3}B_{j}^{2}}{b_{j}\eta_{j}B_{j}}\mathbb{E}\parallel\tilde{x}_{j}-\tilde{x}_{j-1}\parallel^{2}-2\dfrac{\lambda^{3}}{(1-\lambda)}\eta_{j}B_{j}\mathbb{E}\parallel e_{j}\parallel^{2} (\text{~Lemma~\ref{unlemma8}})\\
		&\leq-2(1-\lambda)b_{j}\mathbb{E}<\nabla{f}(\tilde{x}_{j}),(\tilde{x}_{j}-\tilde{x}_{j-1})>+2b_{j}\mathbb{E}(f(\tilde{x}_{j-1})-f(\tilde{x}_{j}))+(2\lambda^{2}L\eta_{j}^{2}B_{j}+2\lambda^{2}\eta_{j}b_{j})\mathbb{E}\parallel e_{j}\parallel^{2}
		\end{aligned}
		\label{unetheorem1}
		\end{equation}
		Using the fact that $2<q,p>\leq\beta\parallel q\parallel^{2}+\dfrac{1}{\beta}\parallel p\parallel^{2}$ for any $\beta>0$, $-2(1-\lambda)b_{j}\mathbb{E}<\nabla{f}(\tilde{x}_{j}), (\tilde{x}_{j}-\tilde{x}_{j-1})>$ in Inq.~\ref{unetheorem1} can be bounded as 
		\begin{equation}
		\begin{aligned}
		&-2(1-\lambda) b_{j}\mathbb{E}<\nabla{f}(\tilde{x}_{j}), (\tilde{x}_{j}-\tilde{x}_{j-1})>\\
		&\leq(1-\lambda)(\dfrac{(1-\lambda) b_{j}\eta_{j}B_{j}}{b_{j}^{3}-(1-\lambda)^{2}\eta_{j}^{2}L^{2}b_{j}B_{j}-(1-\lambda)^{2}\eta_{j}^{3}L^{3}B_{j}^{2}}b_{j}^{2}\mathbb{E}\parallel\nabla{f}(\tilde{x}_{j})\parallel^{2}\\
		&+\dfrac{ b_{j}^{3}-(1-\lambda)^{2}\eta_{j}^{2}L^{2}b_{j}B_{j}-(1-\lambda)^{2}\eta_{j}^{3}L^{3}B_{j}^{2}}{(1-\lambda) b_{j}\eta_{j}B_{j}}\mathbb{E}\parallel \tilde{x}_{j}-\tilde{x}_{j-1}\parallel^{2})
		\end{aligned}
		\end{equation}
		Then Inq.~\ref{unetheorem1} can be expressed as 
		\begin{equation}
		\begin{aligned}
		&\dfrac{\eta_{j}B_{j}}{b_{j}}(2(1-\lambda)-2(1-\lambda)^{2}L\eta_{j}-2(1-\lambda)^{2}\dfrac{b_{j}}{B_{j}}-\dfrac{(1-\lambda)^{2}b_{j}^{3}}{b_{j}^{3}-(1-\lambda)^{2}\eta_{j}^{2}L^{2}b_{j}B_{j}-(1-\lambda)^{2}\eta_{j}^{3}L^{3}B_{j}^{2}})\\
		&\mathbb{E}\parallel\nabla{f}(\tilde{x}_{j})\parallel^{2}\\
		&\leq2\mathbb{E}(f(\tilde{x}_{j-1})-f(\tilde{x}_{j}))+\dfrac{2\eta_{j}B_{j}\lambda^{2}}{b_{j}}(\dfrac{\lambda^{2}}{(1-\lambda)}+\eta_{j}L+\dfrac{b_{j}}{B_{j}})\mathbb{E}\parallel e_{j}\parallel^{2}.
		\end{aligned}   
		\label{unetheorem1.2}
		\end{equation}
		Since $\eta_{j}L=\gamma(\dfrac{b_{j}}{B_{j}})^{\alpha}$, $b_{j}\geq1$ and $B_{j}\geq b_{j}\geq B_{j}^{\beta}$ where $\alpha>0$ and $\beta\geq 0$ by~Eq.~\ref{tu-batchsize}, a one part in left hand side of above inequality can be simplified and positive as following:  
		\begin{equation}
		\label{unb}
		\begin{aligned}
		&b_{j}^{3}-(1-\lambda)^{2}\eta_{j}^{2}L^{2}b_{j}B_{j}-(1-\lambda)^{2}\eta_{j}^{3}L^{3}B_{j}^{2}\\
		&=b_{j}^{3}(1-(1-\lambda)^{2}\gamma^{2}\dfrac{b_{j}^{2\alpha-2}}{B_{j}^{2\alpha-1}}-(1-\lambda)^{2}\gamma^{3}\dfrac{b_{j}^{3\alpha-3}}{B_{j}^{3\alpha-2}})\\
		&\geq b_{j}^{3}(1-(1-\lambda)^{2}\gamma^{2}B_{j}^{-1}-(1-\lambda)^{2}\gamma^{3}B_{j}^{-1})\geq 0.86 b_{j}^{3}
		\end{aligned}
		\end{equation}
		By Eq.\ref{unb}, the left side of Inq.~\ref{unetheorem1.2} can be simplified since the factor of geometry distribution $\gamma\geq 0$ as 
		\begin{equation}
		\begin{aligned}
		&\dfrac{\eta_{j}B_{j}}{b_{j}}(2(1-\lambda)-2(1-\lambda)^{2}L\eta_{j}-2(1-\lambda)^{2}\dfrac{b_{j}}{B_{j}}-\dfrac{(1-\lambda)^{2}b_{j}^{3}}{b_{j}^{3}-(1-\lambda)^{2}\eta_{j}^{2}L^{2}b_{j}B_{j}-(1-\lambda)^{2}\eta_{j}^{3}L^{3}B_{j}^{2}})\\
		&\mathbb{E}\parallel\nabla{f}(\tilde{x}_{j})\parallel^{2}\\
		&\geq\dfrac{\gamma}{L}B_{j}^{\alpha\beta-\alpha-\beta+1}\left(2(1-\lambda)-(2\gamma B_{j}^{\alpha\beta-\alpha}+2\dfrac{b_{j}}{B_{j}})(1-\lambda)^{2}-1.16(1-\lambda)^{2}\right)\mathbb{E}||\nabla{f}(\tilde{x}_{j})||^{2}\\
		&\geq\dfrac{\gamma}{L}B_{j}^{\alpha\beta-\alpha-\beta+1}\left(2(1-\lambda)-(2\gamma+2)B_{j}^{-1}(1-\lambda)^{2}-1.16(1-\lambda)^{2}\right)\mathbb{E}||\nabla{f}(\tilde{x}_{j})||^{2}
		\end{aligned}
		\label{unetheorem1.3}
		\end{equation}
		~Eq.\ref{unetheorem1.3} is positive when $0\leq\gamma\leq\dfrac{13}{50}$ and $B_{j}\geq 3$. Moreover,~\cite{DBLP:conf/nips/LeiJCJ17,pmlr-v54-lei17a} determined the learning rate $\eta=\dfrac{\gamma}{L}\dfrac{b_{j}}{B_{j}}\leq\dfrac{1}{3L}$ that $\gamma\leq\dfrac{1}{3}$ which can guarantees the convergence in non-convex case. In our case, $\gamma\leq\dfrac{13}{50}$ satisfies within the range $\gamma\leq\dfrac{1}{3}$. Then~Eq.\ref{unetheorem1.2} can be simplified by~Eq.\ref{unetheorem1.3} as
		\begin{equation}
		\begin{aligned}
		\mathbb{E}\parallel\nabla{f}(\tilde{x}_{j})\parallel^{2}&\leq\dfrac{2\mathbb{E}[f(\tilde{x}_{j-1})-f(\tilde{x}_{j})]+2\dfrac{\gamma}{L}B_{j}^{\alpha\beta-\alpha-\beta+1}\lambda^{2}(\dfrac{\lambda^{2}}{(1-\lambda)}+B_{j}^{\alpha\beta-\alpha}\gamma+B_{j}^{\beta-\alpha}L)\mathbb{E}|| e_{j}||^{2}}{\dfrac{\gamma}{L}B_{j}^{\alpha\beta-\alpha-\beta+1}\left(2(1-\lambda)-(2\gamma B_{j}^{\alpha\beta-\alpha}+2B_{j}^{\beta-1})(1-\lambda)^{2}-1.16(1-\lambda)^{2}\right)}\\
		&\leq\dfrac{\overbrace{2\mathbb{E}(f(\tilde{x}_{j-1}-f(\tilde{x}_{j})))}^{\text{positive by~Lemma~\ref{lemma2}}}+\overbrace{2\dfrac{\gamma}{L}\lambda^{2}B_{j}^{\alpha\beta-\alpha-\beta+1}B_{j}^{2\alpha}\mathbb{E}\parallel e_{j}\parallel^{2}}^{\text{positive}}}{\dfrac{\gamma}{L}B_{j}^{\alpha\beta-\alpha-\beta+1}\left(2(1-\lambda)-(2\gamma B_{j}^{\alpha\beta-\alpha}+2B_{j}^{\beta-1})(1-\lambda)^{2}-1.16(1-\lambda)^{2}\right)},
		\end{aligned}
		\label{unetheorem1.4}
		\end{equation}
		Then, using~Lemma~\ref{unlemma2}, Inq.~\ref{unetheorem1.4} can be rewritten as 
		\begin{equation}
		\mathbb{E}\parallel\nabla{f}(\tilde{x}_{j})\parallel^{2}
		\leq\dfrac{2\mathbb{E}(f(\tilde{x}_{j-1}-f(\tilde{x}_{j})))+2\dfrac{\gamma}{L}\lambda^{4}B_{j}^{\alpha\beta+\alpha-\beta}I(B_{j}<n)\mathcal{S}^{*}}{\dfrac{\gamma}{L}B_{j}^{\alpha\beta-\alpha-\beta+1}\left(2(1-\lambda)-(2\gamma B_{j}^{\alpha\beta-\alpha}+2B_{j}^{\beta-1})(1-\lambda)^{2}-1.16(1-\lambda)^{2}\right)}.
		\label{unetheorem1.5}
		\end{equation}
		
		Since the learning rate $\eta\leq\dfrac{1}{3L}$ was determined by~\cite{DBLP:conf/nips/LeiJCJ17,pmlr-v54-lei17a} that $\gamma\leq\dfrac{1}{3}$ which guarantees the convergence in non-convex case. Thus $\gamma\leq\dfrac{1}{3}$ as a upper bound is considered in our biased and unbiased cases. Particularly in unbiased cases, $\gamma\leq\dfrac{13}{50}\leq\dfrac{1}{3}$.
	\end{proof}
\end{theorem-non}

\subsection{Biased Estimator Version}
We still provide all useful lemmas we will applied in our proof of theorems at first, and provide a proof sketch for  guidance. For the biased estimation version, we still start by bounding the gradient $\mathbb{E}_{\mathcal{\tilde{I}}_{k}}\parallel v_{k}^{(j)}\parallel^{2}$ in~Lemma~\ref{lemma3} and the variance $\mathbb{E}_{\mathcal{I}_{j}}\parallel e_{j}\parallel^{2}$ in~Lemma~\ref{lemma4}.
\begin{lemma}
	Under~Definition~\ref{lsmooth1}, 
	\begin{equation*}
	\begin{aligned}
	\mathbb{E}_{\mathcal{\tilde{I}}_{k}}\parallel v_{k}^{(j)}\parallel^{2}&\leq \dfrac{(1-\lambda)^{2}L^{2}}{b_{j}}\parallel x_{k}^{(j)}-x_{0}^{(j)}\parallel+2(1-\lambda)^{2}\parallel\nabla f(x_{k}^{(j)})\parallel^{2}+2\parallel e_{j}\parallel^{2}.
	\end{aligned}
	\end{equation*}
	\begin{proof}
		Using the fact that for a random variable $Z$ $\mathbb{E}\parallel Z\parallel^{2}=\mathbb\parallel Z-\mathbb{E}Z\parallel^{2}+\parallel \mathbb{E}Z\parallel^{2}$, we have 
		\begin{equation}\label{bias1}
		\begin{aligned}
		&\mathbb{E}_{\mathcal{\tilde{I}}_{k}}\parallel v_{k}^{(j)}\parallel^{2}=\mathbb{E}_{\mathcal{\tilde{I}}_{k}}\parallel v_{k}^{(j)}-\mathbb{E}_{\mathcal{\tilde{I}}_{k}} v_{k}^{(j)}\parallel^{2}+\parallel\mathbb{E}_{\mathcal{\tilde{I}}_{k}} v_{k}^{(j)}\parallel^{2}\\
		&=\mathbb{E}_{\mathcal{\tilde{I}}_{k}}\parallel(1-\lambda)(\nabla f_{\mathcal{\tilde{I}}_{k}}(x_{k}^{(j)})-\nabla f_{\mathcal{\tilde{I}}_{k}}(x_{0}^{(j)}))-(1-\lambda)(\nabla{f}(x_{k}^{(j)})-\nabla{f}x_{0}^{(j)})\parallel^{2}\\
		&+\parallel(1-\lambda)\nabla{f}(x_{k}^{(j)})+e_{j}\parallel^{2}\\
		&\leq(1-\lambda)^{2}\mathbb{E}_{\mathcal{\tilde{I}}_{k}}\parallel\nabla{f}_{\mathcal{\tilde{I}}_{k}}(x_{k}^{(j)})-\nabla{f}_{\mathcal{\tilde{I}}_{k}}(x_{0}^{(j)})-(\nabla{f}(x_{k}^{(j)})-\nabla{f}x_{0}^{(j)})\parallel^{2}\\
		&+2(1-\lambda)^{2}\parallel\nabla{f}(x_{k}^{(j)})\parallel^{2}+2\parallel e_{j}\parallel^{2}.
		\end{aligned}
		\end{equation}
		By~Lemma~\ref{lemma1}, the first part of inequality in~Eq.\ref{bias1} can be rewritten as,
		\begin{equation}
		\begin{aligned}
		&(1-\lambda)^{2}\mathbb{E}_{\mathcal{\tilde{I}}_{k}}\parallel\nabla{f}_{\mathcal{\tilde{I}}_{k}}(x_{k}^{(j)})-\nabla{f}_{\mathcal{\tilde{I}}_{k}}(x_{0}^{(j)})-(\nabla{f}(x_{k}^{(j)})-\nabla{f}x_{0}^{(j)})\parallel^{2}\\
		&\leq\frac{(1-\lambda)^{2}}{b_{j}}\cdot\frac{1}{n}\sum_{i=1}^{n}\parallel\nabla{f}_{i}(x_{k}^{(j)})-\nabla{f}_{i}(x_{0}^{(j)})-(\nabla{f}(x_{k}^{(j)})-\nabla{f}(x_{0}^{(j)}))\parallel^{2}\\
		&=\frac{(1-\lambda)^{2}}{b_{j}}\cdot(\frac{1}{n}\sum_{i=1}^{n}\parallel\nabla{f}_{i}(x_{k}^{(j)})-\nabla{f}_{i}(x_{0}^{(j)})\parallel^{2}-\parallel(\nabla{f}(x_{k}^{(j)})-\nabla{f}(x_{0}^{(j)}))\parallel^{2})\\
		&\leq\frac{(1-\lambda)^{2}}{b_{j}}\cdot\frac{1}{n}\sum_{i=1}^{n}\parallel\nabla{f}_{i}(x_{k}^{(j)})-\nabla{f}_{i}(x_{0}^{(j)})\parallel^{2}\\
		&\leq\frac{(1-\lambda)^{2}}{b_{j}}\cdot L^{2}\parallel x_{k}^{(j)}-x_{0}^{(j)}\parallel^{2}
		\end{aligned}
		\end{equation}
		where the last line is based on~Definition~\ref{lsmooth1}, then the bound of the gradient can be written as, 
		\begin{equation}
		\begin{aligned}
		\mathbb{E}_{\mathcal{\tilde{I}}_{k}}\parallel v_{k}^{(j)}\parallel^{2}&\leq\frac{(1-\lambda)^{2}L^{2}}{b_{j}}\parallel x_{k}^{(j)}-x_{0}^{(j)}\parallel^{2}+2(1-\lambda)^{2}\parallel\nabla{f}(x_{k}^{(j)})\parallel^{2}+2\parallel e_{j}\parallel^{2}.
		\end{aligned}
		\end{equation}
	\end{proof}
	\label{lemma3}
\end{lemma}

\begin{lemma}
	\begin{equation*}
	\begin{aligned}
	\mathbb{E}_{\mathcal{I}_{j}}\parallel e_{j}\parallel^{2}&\leq(1-\lambda)^{2}\dfrac{I(B_{j}<n)}{B_{j}}\mathcal{S^{*}}+(1-2\lambda)^{2}\mathbb{E}_{\mathcal{I}_{j}}[\nabla{f_{i}}(\tilde{x}_{j-1})]^{2}\\
	&=\mathbb{E}_{\mathcal{I}_{j}}\parallel \tilde{e_{j}}\parallel^{2}+(1-2\lambda)^{2}\mathbb{E}_{\mathcal{I}_{j}}[\nabla{f_{i}}(\tilde{x}_{j-1})]^{2}
	\end{aligned}
	\end{equation*}
	where $(1-\lambda)^{2}\dfrac{I(B_{j}<n)}{B_{j}}\mathcal{S^{*}}=\mathbb{E}_{\mathcal{I}_{j}}\parallel \tilde{e_{j}}\parallel^{2}$ and $0<\lambda<1$.
	\begin{proof}
		Based on~Lemma~\ref{lemma1} and the observation that $\tilde{x}_{j-1}$ is independent of 
		\begin{equation}
\begin{aligned}
     \mathbb{E}_{\mathcal{I}_{j}}\parallel e_{j}\parallel^{2}&=\dfrac{n-B_{j}}{(n-1)B_{j}}\cdot\frac{1}{n}\sum_{i=1}^{n}\parallel(1-\lambda)\nabla{f_{i}}(\tilde{x}_{j-1})-\lambda\nabla{f}(\tilde{x}_{j-1})\parallel^{2}\\
     &=\dfrac{n-B_{j}}{(n-1)B_{j}}\mathbb{E}_{\mathcal{I}_{j}}\parallel(1-\lambda)\nabla{f_{i}}(\tilde{x}_{j-1})-\lambda\mathbb{E}_{\mathcal{I}_{j}}[\nabla{f_{i}}(\tilde{x}_{j-1})]\parallel^{2}\\
     &=\dfrac{n-B_{j}}{(n-1)B_{j}}\mathbb{E}_{\mathcal{I}_{j}}\left[(1-\lambda)^{2}\nabla{f_{i}}(\tilde{x}_{j-1})^{2}-(2\lambda-3\lambda^{2})\mathbb{E}_{\mathcal{I}_{j}}[\nabla{f_{i}}(\tilde{x}_{j-1})]^{2}\right]\\
     &=\dfrac{n-B_{j}}{(n-1)B_{j}}\left[\underbrace{(1-\lambda)^{2}\mathbb{E}_{\mathcal{I}_{j}}\left[\nabla{f_{i}}(\tilde{x}_{j-1})^{2}-\mathbb{E}_{\mathcal{I}_{j}}[\nabla{f_{i}}(\tilde{x}_{j-1})]^{2}\right]}_{Unbiased}+\underbrace{(1-2\lambda)^{2}\mathbb{E}_{\mathcal{I}_{j}}[\nabla{f_{i}}(\tilde{x}_{j-1})]^{2}}_{Extra/ term}\right]\\
&=\dfrac{n-B_{j}}{(n-1)B_{j}}\cdot\left((1-\lambda)^{2}\frac{1}{n}\sum_{i=1}^{n}\parallel\nabla{f_{i}}(\tilde{x}_{j-1})-\nabla{f}(\tilde{x}_{j-1})\parallel^{2}+(1-2\lambda)^{2}\mathbb{E}_{\mathcal{I}_{j}}[\nabla{f_{i}}(\tilde{x}_{j-1})]^{2}\right)\\
&\leq(1-\lambda)^{2}\dfrac{n-B_{j}}{(n-1)B_{j}}\cdot\mathcal{S^{*}}+\dfrac{n-B_{j}}{(n-1)B_{j}}(1-2\lambda)^{2}\mathbb{E}_{\mathcal{I}_{j}}[\nabla{f_{i}}(\tilde{x}_{j-1})]^{2}\\
&\leq(1-\lambda)^{2}\dfrac{I(B_{j}<n)}{B_{j}}\mathcal{S^{*}}+(1-2\lambda)^{2}\mathbb{E}_{\mathcal{I}_{j}}[\nabla{f_{i}}(\tilde{x}_{j-1})]^{2},
\end{aligned}
\label{f38}
\end{equation}
where the upper bound of the variance of the stochastic gradients $\mathcal{S^{*}}=\frac{1}{n}\sum_{i=1}^{n}\parallel\nabla{f_{i}}(\tilde{x}_{j-1})-\nabla{f}(\tilde{x}_{j-1})\parallel^{2}$.In above function, as $\nabla{f}(\tilde{x}_{j-1})$ is the expectation value of $\nabla{f_{i}}(\tilde{x}_{j-1})$, we use $\mathbb{E}_{\mathcal{I}_{j}}[\nabla{f_{i}}(\tilde{x}_{j-1})]$ to alternative $\nabla{f}(\tilde{x}_{j-1})$ for easily understanding later proof. Meanwhile, We can achieve the third equation in above function since the fact that $\mathbb{E}[(1-\lambda)Z-\lambda\mathbb{E}[Z]]^{2}=(1-\lambda)^{2}\mathbb{E}[Z^{2}]-(2\lambda-3\lambda^{2})\mathbb{E}[Z]^{2}=\mathbb{E}[(1-\lambda)^{2}Z^{2}-(2\lambda-3\lambda^{2})\mathbb{E}[Z]^{2}]$.  
\end{proof}
	\label{lemma4}
\end{lemma}

\begin{lemma}
	\label{lemma6} 
	Suppose $\eta_{j}L<1$, then under~Definition~\ref{lsmooth1}, 
	\begin{equation*}
	\begin{aligned}
	&(1-\lambda)(1-(1-\lambda)L\eta_{j})\eta_{j}B_{j}\mathbb{E}\parallel\nabla{f}(\tilde{x}_{j})\parallel^{2}+\eta_{j}B_{j}\mathbb{E}<e_{j},\nabla{f}(\tilde{x}_{j})>\\
	&\leq b_{j}\mathbb{E}(f(\tilde{x}_{j-1})-f(\tilde{x}_{j}))+\dfrac{(1-\lambda)^{2}\eta_{j}^{2}B_{j}L^{3}}{2b_{j}}\mathbb{E}\parallel\tilde{x}_{j}-\tilde{x}_{j-1}\parallel^{2}+L\eta_{j}^{2}B_{j}\mathbb{E}\parallel e_{j}\parallel^{2}.
	\end{aligned}
	\label{elemma6}
	\end{equation*}
	where $\mathbb{E}$ denotes the expectation with respect to all randomness. 
	\begin{proof}
		By~Definition~\ref{lsmooth1}, we have
		\begin{equation}
		\begin{aligned}
		&\mathbb{E}_{\tilde{\mathcal{I}}_{k}}[f(x_{k+1}^{(j)})]\leq f(x_{k}^{(j)})-\eta_{j}<\mathbb{E}_{\tilde{\mathcal{I}}_{k}}v_{k},\nabla{f}(x_{k}^{(j)})>+\dfrac{L\eta_{j}^{2}}{2}\mathbb{E}_{\tilde{\mathcal{I}}_{k}}\parallel v_{k}\parallel^{2}\\
		&=f(x_{k}^{(j)})-\eta_{j}<((1-\lambda)\nabla{f}(x_{k}^{(j)})+e_{j}),\nabla{f}(x)_{k}^{(j)})>+\dfrac{L\eta_{j}^{2}}{2}\mathbb{E}_{\tilde{\mathcal{I}}_{k}}\parallel v_{k}\parallel^{2}\\
		&\leq f(x_{k}^{(j)})-\eta_{j}(1-\lambda)\parallel\nabla{f}(x_{k}^{(j)})\parallel^{2}-\eta_{j}<e_{j},\nabla{f}(x_{k}^{(j)})>\\
		&+\dfrac{L^{3}\eta_{j}^{2}(1-\lambda)^{2}}{2b_{j}}\parallel x_{k}^{(j)}-x_{0}^{(j)}\parallel^{2}+L\eta_{j}^{2}(1-\lambda)^{2}\parallel\nabla{f}(x_{k}^{(j)})\parallel^{2}+L\eta_{j}^{2}\parallel e_{j}\parallel^{2}\\
		&=f(x_{k}^{(j)})-(\eta_{j}(1-\lambda)-L\eta_{j}^{2}(1-\lambda)^{2})\parallel\nabla{f}(x_{k}^{(j)})\parallel^{2}\\
		&-\eta_{j}<e_{j},\nabla{f}(x_{k}^{(j)})>+\dfrac{L^{3}\eta_{j}^{2}(1-\lambda)^{2}}{2b_{j}}\parallel x_{k}^{(j)}-x_{0}^{(j)}\parallel^{2}+L\eta_{j}^{2}\parallel e_{j}\parallel^{2}\\
		\end{aligned}
		\label{le5}
		\end{equation}
		Let $\mathbb{E}_{j}$ denote the expectation $\tilde{\mathcal{I}}_{0}, \tilde{\mathcal{I}}_{1}$,..., given $\tilde{\mathcal{N}}_{j}$ since $\tilde{\mathcal{N}}_{j}$ is independent of them and let k=$\mathcal{N}_{j}$ in Inq~\ref{le5}. As $\tilde{\mathcal{I}}_{k+1}, \tilde{\mathcal{I}}_{k+2}$,... are independent of $x_{k}^{(j)}$ and taking the expectation with respect to $\mathcal{N}_{j}$ and using Fubini's theorem, Inq.~\ref{le5} implies that 
		\begin{equation}
		\begin{aligned}
		&\eta_{j}(1-\lambda)(1-(1-\lambda)L\eta_{j})\mathbb{E}_{\mathcal{N}_{j}}\mathbb{E}_{j}[\parallel\nabla{f}(x_{\mathcal{N}_{j}}^{(j)})\parallel^{2}]+\eta_{j}\mathbb{E}_{\mathcal{N}_{j}}\mathbb{E}_{j}<e_{j},\nabla{f}(x_{\mathcal{N}_{j}}^{(j)})>\\
		&\leq\mathbb{E}_{\mathcal{N}_{j}}(\mathbb{E}_{j}[f(x_{\mathcal{N}_{j}}^{(j)})]-\mathbb{E}_{j}[f(x_{\mathcal{N}_{j+1}}^{(j)})])+\dfrac{L^{3}\eta_{j}^{2}(1-\lambda)^{2}}{2b_{j}}\mathbb{E}_{\mathcal{N}_{j}}\mathbb{E}_{j}\mathbb{E}[\parallel x_{\mathcal{N}_{j}}^{(j)}-x_{0}^{(j)}\parallel^{2}]+L\eta_{j}^{2}\parallel e_{j}\parallel^{2}\\
		&=\dfrac{b_{j}}{B_{j}}(f(x_{0}^{(j)})-\mathbb{E}_{j}\mathbb{E}_{\mathcal{N}_{j}}[f_{\mathcal{N}_{j}}^{(j)}])+\dfrac{L^{3}\eta_{j}^{2}(1-\lambda)^{2}}{2b_{j}}\mathbb{E}_{j}\mathbb{E}_{\mathcal{N}_{j}}[\parallel x_{\mathcal{N}_{j}}^{(j)}-x_{0}^{(j)}\parallel^{2}]+L\eta_{j}^{2}\parallel e_{j}\parallel^{2}
		\end{aligned}  
		\label{le6}
		\end{equation}
		where the last equation in Inq.~\ref{le6} follows from~Lemma~\ref{lemma2}. The lemma substitutes $x_{\mathcal{N}_{j}}^{(j)}(x_{0}^{j})$ by $\tilde{x}_{j}(\tilde{x}_{j-1})$. 
	\end{proof}
\end{lemma}	

\begin{lemma}
	\label{lemma7}
	Suppose $\eta_{j}^{2}L^{2}B_{j}<b_{j}^{2}$, then under Definition lsmooth1, 
	\begin{equation*}
	\begin{aligned}
	&(b_{j}-\dfrac{(1-\lambda)^{2}\eta_{j}^{2}L^{2}B_{j}}{b_{j}})\mathbb{E}[\parallel\tilde{x}_{j}-\tilde{x}_{j-1}\parallel^{2}]+2\eta_{j}B_{j}\mathbb{E}<e_{j},(\tilde{x}_{j}-\tilde{x}_{j-1})>\\
	&\leq-2(1-\lambda)\eta_{j}B_{j}\mathbb{E}<\nabla{f}(\tilde{x}_{j}),(\tilde{x}_{j}-\tilde{x}_{j-1})>+2(1-\lambda)^{2}\eta_{j}^{2}B_{j}\mathbb{E}[\parallel\nabla{f}(\tilde{x}_{j})\parallel^{2}]+2\eta_{j}^{2}B_{j}\mathbb{E}[\parallel e_{j}\parallel^{2}]
	\end{aligned}  
	\label{elemma7}
	\end{equation*}
	\begin{proof}
		Since $x_{k+1}^{(j)}=x_{k}^{(j)}-\eta_{j}v_{k}^{(j)}$, we have 
		\begin{equation}
		\begin{aligned}
		&\mathbb{E}_{\mathcal{\tilde{I}}_{k}}[\parallel x_{k+1}^{(j)}-x_{0}^{(j)}\parallel^{2}]\\
		&=\parallel x_{k}^{(j)}-x_{0}^{(j)}\parallel^{2}-2\eta_{j}<\mathbb{E}_{\mathcal{\tilde{I}}_{k}}v_{k}^{(j)},(x_{k}^{(j)}-x_{0}^{(j)})>+\eta^{2}_{j}\mathbb{E}_{\mathcal{\tilde{I}}_{k}}\parallel v_{k}^{(j)}\parallel^{2}\\
		&=\parallel x_{k}^{(j)}-x_{0}^{(j)}\parallel^{2}-2\eta_{j}(1-\lambda)<\nabla{f}(x_{k}^{(j)}),(x_{k}^{(j)}-x_{0}^{(j)})>-2\eta_{j}<e_{j},(x_{k}^{(j)}-x_{0}^{(j)})>+\eta_{j}^{2}\mathbb{E}_{\mathcal{\tilde{I}}_{k}}\parallel v_{k}^{(j)}\parallel^{2}\\
		&\leq(1+\dfrac{(1-\lambda)^{2}\eta_{j}^{2}L^{2}}{b_{j}})\parallel x_{k}^{(j)}-x_{0}^{(j)}\parallel^{2}-2\eta_{j}(1-\lambda)<\nabla{f}(x_{k}^{(j)}), x_{k}^{(j)}-x_{0}^{(j)}>-2\eta_{j}<e_{j},(x_{k}^{(j)}-x_{0}^{(j)})>\\
		&+2(1-\lambda)^{2}\eta_{j}^{2}\parallel\nabla{f}(x_{k}^{(j)})\parallel^{2}+2\eta_{j}^{2}\parallel e_{j}\parallel^{2}.
		\end{aligned}
		\end{equation}
		where the last inequality is based on~Lemma~\ref{lemma3}. Using the same notation $\mathbb{E}_{j}$ in~Eq.~\ref{tu-batchsize} we have
		\begin{equation}
		\begin{aligned}
		&2\eta_{j}(1-\lambda)\mathbb{E}_{j}<\nabla{f}(x_{k}^{(j)}),(x_{k}^{(j)}-x_{0}^{(j)})>+2\eta_{j}\mathbb{E}_{j}<e_{j},(x_{k}^{(j)}-x_{0}^{(j)})>\\
		&\leq(1+\dfrac{(1-\lambda)^{2}\eta_{j}^{2}L^{2}}{b_{j}})\mathbb{E}_{j}\parallel x_{k}^{(j)}-x_{0}^{(j)}\parallel^{2}-\mathbb{E}_{j}\parallel x_{k+1}^{(j)}-x_{0}^{(j)}\parallel^{2}+2(1-\lambda)^{2}\eta_{j}^{2}\parallel \nabla{f}(x_{k}^{(j)})\parallel^{2}+2\eta_{j}^{2}\parallel e_{j}\parallel^{2}
		\end{aligned}
		\end{equation}
		Let $k=N_{j}$, and using Fubini's theorem, we have, 
		\begin{equation}
		\begin{aligned}
		&2\eta_{j}(1-\lambda)\mathbb{E}_{N_{j}}\mathbb{E}_{j}<\nabla{f}(x_{N_{j}}^{(j)}),(x_{N_{j}}^{(j)}-x_{0}^{(j)})>+2\eta_{j}\mathbb{E}_{N_{j}}\mathbb{E}_{j}<e_{j},(x_{N_{j}}^{(j)}-x_{0}^{(j)})>\\
		&\leq(1+\dfrac{(1-\lambda)^{2}\eta_{j}L^{2}}{b_{j}})\mathbb{E}_{N_{j}}\mathbb{E}_{j}\parallel x_{N_{j}}^{(j)}-x_{0}^{(j)}\parallel^{2}-\mathbb{E}_{N_{j}}\mathbb{E}_{j}\parallel x_{N_{j}+1}^{(j)}-x_{0}^{(j)}\parallel^{2}\\
		&+2(1-\lambda)^{2}\eta_{j}^{2}\mathbb{E}_{N_{j}}\parallel\nabla{f}(x_{N_{j}}^{(j)})\parallel^{2}+2\eta_{j}^{2}\parallel e_{j}\parallel^{2}\\
		&=(-\dfrac{b_{j}}{B_{j}}+\dfrac{(1-\lambda)^{2}\eta_{j}^{2}L^{2}}{b_{j}})\mathbb{E}_{N_{j}}\mathbb{E}_{j}\parallel x_{N_{j}}^{(j)}-x_{0}^{(j)}\parallel^{2}+2(1-\lambda)^{2}\eta^{2}_{j}\mathbb{E}_{N_{j}}\parallel\nabla{f}(x_{N_{j}}^{(j)})\parallel^{2}+2\eta_{j}^{2}\parallel e_{j}\parallel^{2}.
		\end{aligned}
		\end{equation}
		The lemma is then proved by substituting $x_{N_{j}}^{(j)}(x_{0}^{(j)})$ by $\tilde{x}_{j}(\tilde{x}_{j-1})$.
	\end{proof}
\end{lemma}
\begin{lemma}
	\label{lemma8}
	\begin{equation*}
	\begin{aligned}
	&b_{j}\mathbb{E}<e_{j},(\tilde{x}_{j}-\tilde{x}_{j-1})>=-\eta_{j}(1-\lambda)B_{j}\mathbb{E}<e_{j},\nabla{f}(\tilde{x}_{j})>-\eta_{j}B_{j}\mathbb{E}\parallel e_{j}\parallel^{2}
	\end{aligned}
	\end{equation*}
	\begin{proof}
		Let $M_{k}^{(j)}=<e_{j},(x_{k}^{(j)}-x_{0}^{(j)})>$, then we have
		\begin{equation*}
		\mathbb{E}_{N_{j}}<e_{j},(\tilde{x}_{j}-\tilde{x}_{j-1})>=\mathbb{E}_{N_{j}}M_{N_{j}}^{(j)}.
		\end{equation*}
		Since $N_{j}$ is independent of $(x_{0}^{(j)}, e_{j})$, it has 
		\begin{equation}
		\mathbb{E}<e_{j},(\tilde{x}_{j}-\tilde{x}_{j-1})>=\mathbb{E}M_{N_{j}}^{(j)}.
		\end{equation}
		Also $M_{0}^{(j)}=0$, then we have 
		\begin{equation}
		\begin{aligned}
		&\mathbb{E}_{\mathcal{\tilde{I}}_{k}}(M_{k+1}^{(j)}-M_{k}^{(j)})\\
		&=\mathbb{E}_{\mathcal{\tilde{I}}_{k}}<e_{j},(x_{k+1}^{(j)}-x_{k}^{(j)})>=-\eta_{j}<e_{j},\mathbb{E}_{\mathcal{\tilde{I}}_{k}}[v_{k}^{(j)}]>\\
		&=-\eta_{j}(1-\lambda)<e_{j},\nabla{f}(x_{k}^{(j)})>-\eta_{j}\parallel e_{j}\parallel^{2}.
		\end{aligned}
		\end{equation}
		Using the same notation $\mathbb{E}_{j}$ in~Eq.~\ref{tu-batchsize}, we have
		\begin{equation}
		\mathbb{E}_{j}(M_{k+1}^{(j)}-M_{k}^{(j)})=-\eta_{j}(1-\lambda)<e_{j},\mathbb{E}_{j}\nabla{f}(x_{k}^{(j)})>-\eta_{j}\parallel e_{j}\parallel^{2}.
		\label{elemma8}
		\end{equation}
		Let $k=N_{j}$ in~Eq.\ref{elemma8}. Using Fubini's theorem and~Lemma~\ref{lemma2}, we have,
		\begin{equation}
		\dfrac{b_{j}}{B_{j}}\mathbb{E}_{N_{j}}M_{N_{j}}^{(j)}=-\eta_{j}(1-\lambda)<e_{j},\mathbb{E}_{N_{j}}\mathbb{E}_{j}\nabla{f}(x_{k}^{(j)})>-\eta_{j}\parallel e_{j}\parallel^{2}.
		\end{equation}
		The lemma is then proved by substituting $x_{N_{j}}^{(j)}(x_{0}^{(j)})$ by $\tilde{x}_{j}(\tilde{x}_{j-1})$.
	\end{proof}
\end{lemma}	
\section*{Proof of~Theorem~\ref{t1}}
\begin{theorem-non}
	let $\eta_{j}L=\gamma(\dfrac{b_{j}}{B_{j}})^{\alpha}$ $(0\leq\alpha\leq 1)$ and $0\leq\gamma\leq \dfrac{1}{3}$. Suppose $B_{j}\geq b_{j}\geq B_{j}^{\beta}$ $(0\leq\beta\leq 1)$ for all $j$, then under~Definition~\ref{lsmooth1}, the output $\tilde{x}_{j}$ of Alg~\ref{a6.2} we have,
\begin{equation*}
\begin{aligned}
&\mathbb{E}\|\nabla{f}(\tilde{x}_{j})\|^{2}\leq\dfrac{(\dfrac{2L}{\gamma})\triangle_{f}}{{\Theta\sum_{j=1}^{T}b_{j}^{\alpha-1}B_{j}^{1-\alpha}}}+\dfrac{2(1-\lambda)^{2}I(B_{j}<n))\mathcal{S}^{*}}{\Theta B_{j}^{1-2\alpha}},
\end{aligned}
\end{equation*}
where $0<\lambda<1$ and $\Theta=2(1-\lambda)-(2\gamma B_{j}^{\alpha\beta-\alpha}+2B_{j}^{\beta-1}-4LB_{j}^{2\alpha-2})(1-\lambda)^{2}-1.16(1-\lambda)^{2}$. 
	\begin{tcolorbox}
	\textit{Proof Sketch}: Combine two equations in Lemma~\ref{lemma6} and Lemma~\ref{lemma7}, we can achieve a upper bound of biased version gradient in single epoch. And further use Lemma~\ref{lemma4}, the final result of Theorem~\ref{t1} can be achieved. 
\end{tcolorbox}
	\begin{proof}
		Multiplying~Eq.\ref{elemma6} by 2 and~Eq.\ref{elemma7} by $\dfrac{b_{j}}{\eta_{j}B_{j}}$ and summing them, then we have,
		\begin{equation}
		\begin{aligned}
		&2\eta_{j}B_{j}(1-\lambda)(1-(1-\lambda)L\eta_{j}-\dfrac{(1-\lambda)b_{j}}{B_{j}})\mathbb{E}\parallel\nabla{f}(\tilde{x}_{j})\parallel^{2}\\
		&+\dfrac{b_{j}^{3}-(1-\lambda)^{2}\eta_{j}^{2}L^{2}b_{j}B_{j}-(1-\lambda)^{2}\eta_{j}^{3}L^{3}B_{j}^{2}}{b_{j}\eta_{j}B_{j}}\mathbb{E}\parallel\tilde{x}_{j}-\tilde{x}_{j-1}\parallel^{2}\\
		&+2\eta_{j}B_{j}\mathbb{E}<e_{j},\nabla{f}(\tilde{x}_{j})>+2b_{j}\mathbb{E}<e_{j},(\tilde{x}_{j}-\tilde{x}_{j-1})>\\
		&=2\eta_{j}B_{j}(1-\lambda)(1-(1-\lambda)L\eta_{j}-\dfrac{(1-\lambda)b_{j}}{B_{j}}+\dfrac{(2\lambda-1)^{2}}{2\eta_{j}B_{j}(1-\lambda)})\mathbb{E}\parallel\nabla{f}(\tilde{x}_{j})\parallel^{2}\\
		&+\dfrac{b_{j}^{3}-(1-\lambda)^{2}\eta_{j}^{2}L^{2}b_{j}B_{j}-(1-\lambda)^{2}\eta_{j}^{3}L^{3}B_{j}^{2}}{b_{j}\eta_{j}B_{j}}\mathbb{E}\parallel\tilde{x}_{j}-\tilde{x}_{j-1}\parallel^{2}-2\eta_{j}B_{j}\mathbb{E}\parallel \tilde{e_{j}}\parallel^{2} (\text{~Lemma~\ref{lemma8}})\\
		&\leq-2(1-\lambda)b_{j}\mathbb{E}<\nabla{f}(\tilde{x}_{j}),(\tilde{x}_{j}-\tilde{x}_{j-1})>+2b_{j}\mathbb{E}(f(\tilde{x}_{j-1})-f(\tilde{x}_{j}))+(2L\eta_{j}^{2}B_{j}+2\eta_{j}b_{j})\mathbb{E}\parallel \tilde{e_{j}}\parallel^{2}
		\end{aligned}
		\label{etheorem1}
		\end{equation}
		Using the fact that $2<q,p>\leq\beta\parallel q\parallel^{2}+\dfrac{1}{\beta}\parallel p\parallel^{2}$ for any $\beta>0$, $-2b_{j}\mathbb{E}<\nabla{f}(\tilde{x}_{j}), (\tilde{x}_{j}-\tilde{x}_{j-1})>$ in Inq.~\ref{etheorem1} can be bounded as 
		\begin{equation}
		\begin{aligned}
		&-2(1-\lambda)b_{j}\mathbb{E}<\nabla{f}(\tilde{x}_{j}), (\tilde{x}_{j}-\tilde{x}_{j-1})>\\
		&\leq(1-\lambda)(\dfrac{(1-\lambda)b_{j}\eta_{j}B_{j}}{b_{j}^{3}-(1-\lambda)^{2}\eta_{j}^{2}L^{2}b_{j}B_{j}-(1-\lambda)^{2}\eta_{j}^{3}L^{3}B_{j}^{2}}b_{j}^{2}\mathbb{E}\parallel\nabla{f}(\tilde{x}_{j})\parallel^{2}\\
		&+\dfrac{b_{j}^{3}-(1-\lambda)^{2}\eta_{j}^{2}L^{2}b_{j}B_{j}-(1-\lambda)^{2}\eta_{j}^{3}L^{3}B_{j}^{2}}{(1-\lambda)b_{j}\eta_{j}B_{j}}\mathbb{E}\parallel \tilde{x}_{j}-\tilde{x}_{j-1}\parallel^{2})
		\end{aligned}
		\end{equation}
		Then Inq.~\ref{etheorem1} can be rewritten as 
		\begin{equation}
		\begin{aligned}
		&\dfrac{\eta_{j}B_{j}}{b_{j}}(2(1-\lambda)-2(1-\lambda)^{2}L\eta_{j}-2(1-\lambda)^{2}\dfrac{b_{j}}{B_{j}}+\dfrac{(2\lambda-1)^{2}}{\eta_{j}B_{j}}\\
		&-\dfrac{(1-\lambda)^{2}b_{j}^{3}}{b_{j}^{3}-(1-\lambda)^{2}\eta_{j}^{2}L^{2}b_{j}B_{j}-(1-\lambda)^{2}\eta_{j}^{3}L^{3}B_{j}^{2}})\mathbb{E}\parallel\nabla{f}(\tilde{x}_{j})\parallel^{2}\\
		&\leq2\mathbb{E}(f(\tilde{x}_{j-1})-f(\tilde{x}_{j}))+\dfrac{2\eta_{j}B_{j}}{b_{j}}(1+\eta_{j}L+\dfrac{b_{j}}{B_{j}})\mathbb{E}\parallel \tilde{e_{j}}\parallel^{2}.
		\end{aligned}   
		\label{etheorem1.2}
		\end{equation}
		Since $\eta_{j}L=\gamma(\dfrac{b_{j}}{B_{j}})^{\alpha}$, $b_{j}\geq1$ and $B_{j}\geq b_{j}\geq B_{j}^{\beta}$ where $0<\alpha\leq1, 0\leq \beta\leq1$, we have 
		\begin{equation}
		\label{b3}
		\begin{aligned}
		&b_{j}^{3}-(1-\lambda)^{2}\eta_{j}^{2}L^{2}b_{j}B_{j}-(1-\lambda)^{2}\eta_{j}^{3}L^{3}B_{j}^{2}\\
		&=b_{j}^{3}(1-(1-\lambda)^{2}\gamma^{2}\dfrac{b_{j}^{2\alpha-2}}{B_{j}^{2\alpha-1}}-(1-\lambda)^{2}\gamma^{3}\dfrac{b_{j}^{3\alpha-3}}{B_{j}^{3\alpha-2}})\\
		&=b_{j}^{3}(1-(1-\lambda)^{2}\gamma^{2}B_{j}^{-1}-(1-\lambda)^{2}\gamma^{3}B_{j}^{-1})\geq 0.86 b_{j}^{3}
		\end{aligned}
		\end{equation}
		By Eq.~\ref{b3}, the left side of Inq.~\ref{etheorem1.2} can be simplified as
		\begin{equation}
		\begin{aligned}
		&\dfrac{\eta_{j}B_{j}}{b_{j}}(2(1-\lambda)-2(1-\lambda)^{2}L\eta_{j}-2(1-\lambda)^{2}\dfrac{b_{j}}{B_{j}}+\dfrac{(2\lambda-1)^{2}}{\eta_{j}B_{j}}-\dfrac{(1-\lambda)^{2}b_{j}^{3}}{b_{j}^{3}-\eta_{j}^{2}L^{2}b_{j}B_{j}-\eta_{j}^{3}L^{3}B_{j}^{2}})\mathbb{E}\parallel\nabla{f}(\tilde{x}_{j})\parallel^{2}\\
		&=\dfrac{\gamma}{L}B_{j}^{1-\alpha+\alpha\beta-\beta}\left(2(1-\lambda)-(2\gamma B_{j}^{\alpha\beta-\alpha}+2B_{j}^{\beta-1})(1-\lambda)^{2}+\dfrac{(2\lambda-1)^{2}}{\dfrac{\gamma}{L}B_{j}^{2\alpha-2}}-1.16(1-\lambda)^{2}\right)\mathbb{E}\parallel\nabla{f}(\tilde{x}_{j})\parallel^{2}\\
		&\geq\dfrac{\gamma}{L}B_{j}^{\alpha\beta-\alpha-\beta+1}\left(2(1-\lambda)-(2\gamma B_{j}^{-1}+2B_{j}^{-1}-4)(1-\lambda)^{2}-1.16(1-\lambda)^{2}\right)\mathbb{E}\parallel\nabla{f}(\tilde{x}_{j})\parallel^{2}.
		\end{aligned}
		\label{etheorem1.3}
		\end{equation}
		~Eq.\ref{etheorem1.3} is positive when $0\leq\gamma\leq 2.42B_{j}-1$ and $B_{j}\geq 1$. Moreover,~\cite{DBLP:conf/nips/LeiJCJ17,pmlr-v54-lei17a} determined the learning rate $\eta=\dfrac{\gamma}{L}\dfrac{b_{j}}{B_{j}}\leq\dfrac{1}{3L}$ that $\gamma\leq\dfrac{1}{3}$ which can guarantees the convergence in non-convex case. In our case, $\gamma$ should satisfy the range $0\leq\gamma\leq\dfrac{1}{3}\leq 2.42B_{j}-1$, thus $\gamma\leq\dfrac{1}{3}$.		
		
		Then~Eq.\ref{etheorem1.2} can be simplified by~Eq.\ref{etheorem1.3} as
		\begin{equation}
		\begin{aligned}
		&\mathbb{E}\parallel\nabla{f}(\tilde{x}_{j})\parallel^{2}\leq\dfrac{2\mathbb{E}[f(\tilde{x}_{j-1})-f(\tilde{x}_{j})]+2\dfrac{\gamma}{L}B_{j}^{\alpha\beta-\alpha-\beta+1}(1+B_{j}^{\alpha\beta-\alpha}\gamma+B_{j}^{b-a}L)\mathbb{E}\parallel e_{j}\parallel^{2}}{\dfrac{\gamma}{L}B_{j}^{1-\alpha+\alpha\beta-\beta}\left(2(1-\lambda)-(2\gamma B_{j}^{\alpha\beta-\alpha}+2B_{j}^{\beta-1}-4LB_{j}^{2\alpha-2})(1-\lambda)^{2}-1.16(1-\lambda)^{2}\right)}\\
		&\leq\dfrac{\overbrace{2\mathbb{E}[f(\tilde{x}_{j-1})-f(\tilde{x}_{j})]}^{\text{positive by~Lemma~\ref{lemma2}}}+\overbrace{2\dfrac{\gamma}{L}B_{j}^{\alpha\beta-\alpha-\beta+1}B_{j}^{2a}\mathbb{E}\parallel e_{j}\parallel^{2}}^{\text{positive}}}{\dfrac{\gamma}{L}B_{j}^{1-\alpha+\alpha\beta-\beta}\left(2(1-\lambda)-(2\gamma B_{j}^{\alpha\beta-\alpha}+2B_{j}^{\beta-1}-4LB_{j}^{2\alpha-2})(1-\lambda)^{2}-1.16(1-\lambda)^{2}\right)}.
		\end{aligned}
		\label{etheorem1.4}
		\end{equation}
        Then, using~Lemma~\ref{lemma4}, Inq.~\ref{etheorem1.4} can be expressed as 
		\begin{equation}
		\begin{aligned}
		\mathbb{E}\parallel\nabla{f}(\tilde{x}_{j})\parallel^{2}&\leq
		\dfrac{2\mathbb{E}[f(\tilde{x}_{j-1})-f(\tilde{x}_{j})]+2(1-\lambda)^{2}\dfrac{\gamma}{L}B_{j}^{\alpha\beta+\alpha-\beta}I(B_{j}<n)\mathcal{S}^{*}}{\dfrac{\gamma}{L}B_{j}^{1-\alpha+\alpha\beta-\beta}\left(2(1-\lambda)-(2\gamma B_{j}^{\alpha\beta-\alpha}+2B_{j}^{\beta-1}-4LB_{j}^{2\alpha-2})(1-\lambda)^{2}-1.16(1-\lambda)^{2}\right)}\\
		&=\dfrac{(\frac{2L}{\gamma})(\frac{b_{j}}{B_{j}})^{1-\alpha}\mathbb{E}(f(\tilde{x}_{j-1})-f(\tilde{x}_{j}))+2(1-\lambda)^{2}\frac{I(B_{j}<n))}{B_{j}^{1-2\alpha}}\mathcal{S}^{*}}{2(1-\lambda)-(2\gamma B_{j}^{\alpha\beta-\alpha}+2B_{j}^{\beta-1}-4LB_{j}^{2\alpha-2})(1-\lambda)^{2}-1.16(1-\lambda)^{2}}
		\end{aligned}
		\label{etheorem1.5}
		\end{equation}
	\end{proof}
\end{theorem-non}
\section{Convergence Analysis for L-smooth Objectives}
	Under the specifications of~Theorem~\ref{t1u}, Theorem~\ref{t1} and~Definition~\ref{lsmooth1},  the output $\tilde{x}_{T}^{*}$ can achieve its upper bound of gradients depending on two estimators.  
	\begin{itemize}
		\item For the unbiased estimator (Alg.~\ref{a3}), $0<\lambda<1$. The upper bound is given by,
		\begin{equation*}
		\mathbb{E}\parallel\nabla{f}(\tilde{x}_{T}^{*})\parallel^{2}
		\leq\dfrac{(\dfrac{2L}{\gamma})\triangle_{f}}{{\theta\sum_{j=1}^{T}b_{j}^{\alpha-1}B_{j}^{1-\alpha}}}+\dfrac{2\lambda^{4}I(B_{j}<n))\mathcal{S}^{*}}{\theta B_{j}^{1-2\alpha}},
		\end{equation*}
		\item For the biased estimator (Alg.~\ref{a6.2}), $0<\lambda<1$. The upper bound is shown as,
		\begin{equation*}
		\mathbb{E}\parallel\nabla{f}(\tilde{x}_{j})\parallel^{2}\leq\dfrac{(\dfrac{2L}{\gamma})\triangle_{f}}{{\Theta\sum_{j=1}^{T}b_{j}^{\alpha-1}B_{j}^{1-\alpha}}}+\dfrac{2(1-\lambda)^{2}I(B_{j}<n))\mathcal{S}^{*}}{\Theta B_{j}^{1-2\alpha}},
		\end{equation*}
	\end{itemize}

	\begin{proof}
		Since $\tilde{x}^{*}_{T}$ is a random element from $(\tilde{x}_{j})_{j=1}^{T}$ with 
		\begin{equation}
		P(\tilde{x}^{*}_{T}=\tilde{x}_{j})\propto\dfrac{\eta_{j}B_{j}}{b_{j}}\propto(\dfrac{B_{j}}{b_{j}})^{\alpha},
		\end{equation}
		Inq.~\ref{unetheorem1.5} and~\ref{etheorem1.5} will be re-scaled as Inq.~\ref{unetheorem2} and \ref{etheorem2} respectively. 
		\begin{itemize}
			\item For the unbiased estimator (Alg.~\ref{a3}), the upper bound is shown as,
			\begin{equation}\label{unetheorem2}
			\mathbb{E}\parallel\nabla{f}(\tilde{x}_{T}^{*})\parallel^{2}
			\leq\dfrac{(\dfrac{2L}{\gamma})\triangle_{f}}{{\theta\sum_{j=1}^{T}b_{j}^{\alpha-1}B_{j}^{1-\alpha}}}+\dfrac{2\lambda^{4}I(B_{j}<n))\mathcal{S}^{*}}{\theta B_{j}^{1-2\alpha}},
			\end{equation}
			where $\theta=2(1-\lambda)-(2\gamma B_{j}^{\alpha\beta-\alpha}+2B_{j}^{\beta-1})(1-\lambda)^{2}-1.16\lambda^{2}$.
			\item For the biased estimator (Alg.~\ref{a6.2}), the upper bound is shown as,
			\begin{equation}\label{etheorem2}
			\mathbb{E}\parallel\nabla{f}(\tilde{x}_{j})\parallel^{2}\leq
			\dfrac{(\dfrac{2L}{\gamma})\triangle_{f}}{{\Theta\sum_{j=1}^{T}b_{j}^{\alpha-1}B_{j}^{1-\alpha}}}+\dfrac{(1-\lambda)^{2}I(B_{j}<n))\mathcal{S}^{*}}{\Theta B_{j}^{1-2\alpha}},
			\end{equation}
			\label{all}
			where $\Theta=2(1-\lambda)-(2\gamma B_{j}^{\alpha\beta-\alpha}+2B_{j}^{\beta-1}-4LB_{j}^{2\alpha-2})(1-\lambda)^{2}-1.16(1-\lambda)^{2}$.
		\end{itemize}
	\end{proof}
After achieved the result in above, and specified parameters, we can obtain result of Theorem~\ref{vcsg}.

\bibliographystyle{unsrt}  
\bibliography{acmart.bib}  


\end{document}